\documentclass{sig-alternate-05-2015}

\usepackage{amsmath}
\usepackage{amsfonts}
\usepackage{graphicx}
\usepackage{array}

\usepackage[caption=false]{subfig}
\usepackage{algorithm,algpseudocode}
\algnewcommand\algorithmicinput{\textbf{Input:}}
\algnewcommand\INPUT{\item[\algorithmicinput]}

\algnewcommand{\Initialize}[1]{%
  \State \textbf{Initialize:}
  \Statex \hspace*{\algorithmicindent}\parbox[t]{.8\linewidth}{\raggedright #1}
}

\algnewcommand\algorithmicoutput{\textbf{Output}}
\algnewcommand\OUTPUT{\item[\algorithmicoutput]}

\algnewcommand{\Output}[1]{%
  \State \textbf{Output:}
  \Statex \hspace*{\algorithmicindent}\parbox[t]{.8\linewidth}{\raggedright #1}
}

\newcommand{\algrule}[1][.2pt]{\par\vskip.5\baselineskip\hrule height #1\par\vskip.5\baselineskip}
\usepackage{tikz}

\begin{document}

% Copyright
%\setcopyright{acmcopyright}

% DOI
%\doi{10.475/123_4}

% ISBN
%\isbn{123-4567-24-567/08/06}

%Conference
%\conferenceinfo{PLDI '13}{June 16--19, 2013, Seattle, WA, USA}

%\acmPrice{\$15.00}

%
% --- Author Metadata here ---
%\conferenceinfo{WOODSTOCK}{'97 El Paso, Texas USA}
% --- End of Author Metadata ---

\title{Inertial Regularization and Selection (IRS)\textemdash{}Sequential Regression in High-Dimension and Sparsity}
%********\\subtitle{[Extended Abstract]
%********\\titlenote{A full version of this paper is available as
%********\\textit{Author's Guide to Preparing ACM SIG Proceedings Using
%********\\LaTeX$2_\epsilon$\ and BibTeX} at
%********\\texttt{www.acm.org/eaddress.htm}}}


\numberofauthors{3} %  in this sample file, there are a *total*
\author{
% 1st. author
\alignauthor
Chitta Ranjan\\
       \affaddr{Georgia Institute of Technology}\\
       \affaddr{Atlanta, U.S.A.}\\
       \email{\texttt{nk.chitta.ranjan@gatech.edu}}
% 2nd. author
\alignauthor
Samaneh Ebrahimi\\
       \affaddr{Georgia Institute of Technology}\\
       \affaddr{Atlanta, U.S.A.}\\
       \email{\texttt{samaneh.ebrahimi@gatech.edu}}
% 3rd. author
\alignauthor Kamran Paynabar\\
       \affaddr{Georgia Institute of Technology}\\
       \affaddr{Atlanta, U.S.A.}\\
       \email{\texttt{kamran.paynabar@gatech.edu}}
}

\maketitle
\begin{abstract}
In this paper, we develop a new sequential regression modeling approach
for data streams. Data streams are commonly found around us, e.g in
a retail enterprise sales data is continuously collected every day.
A demand forecasting model is an important outcome from the data that
needs to be continuously updated with the new incoming data. The main
challenge in such modeling arises when there is a) high dimensional and sparsity, b) need for an adaptive use of prior knowledge, and/or c) structural
changes in the system. The proposed approach addresses these challenges
by incorporating an adaptive L1-penalty and \emph{inertia}
terms in the loss function, and thus called Inertial Regularization
and Selection (IRS). The former term performs model selection to handle
the first challenge while the latter is shown to address the last
two challenges. A recursive estimation algorithm is developed, and
shown to outperform the commonly used state-space models, such as
Kalman Filters, in experimental studies and real data.
\end{abstract}




%\ccsdesc[500]{Computer systems organization~Embedded systems}


%
%  Use this command to print the description
%

% We no longer use \terms command


\section{Introduction}

In recent times, huge amount of data is collected every day for most
of statistical learning processes. For eg., Twitter, Facebook, Walmart,
Weather, generates\textbf{ }tremendous\textbf{ }amount of data everyday.
This leads to accumulation of large amounts of data over time. Such
processes are also sometimes called as a data-streaming process. Modeling
an entire datastream together is computationally challenging. Although,
methods like Stochastic Gradient Method has been proven to work on
such large data, it will fail when the process is also changing over
time. We call such changing systems as evolving. 

Take for example, the behavior of users on YouTube evolves over time.
The user usage data is collected by YouTube, which should be used
for continuously modeling and updating their recommender system. Another
example can be found at retailers demand forecasting models. For eg.,
at Walmart several thousands of products are sold. For each type,
several new brands are introduced or replaced every day. The demands
of all these products interact with each other. The model should,
thus, be continuously updated to accomodate for the dynamic changes
made in the demand system.

Sequential modeling of such processes can alleviate the challenges
posed by the dynamic changes in evolution in the system. Moreover,
it significantly reduces the computation complexities. Several researchers
have worked in this direction to develop methods specific to certain
problems, for eg. evolutionary recommendation systems \cite{ding2005}.

Broadly, the existing methods use a data-pruning or sequential models.
A data-pruning approach takes the most recent data, either using a
moving window or giving decreasing weight based on the oldness of
the data. These methods are, however, ad-hoc with the window size
or the weight functions difficult to optimize. Moreover, they do not
appropriately use the model information learnt so far.

Sequential models, on the other hand, are better at using the past
information on the model and updating it from the newly observed data.
However, the commonly used sequential methods, like Kalman Filters,
Particle Filter, etc. are not suitable for high-dimensional data.
Moreover, they are not capable of optimally selecting a sub-model
when there is sparsity.

On the other hand, the abovementioned problems are high-dimensional
and sparse. Besides, following are the inherent challenges in the
problem at hand,
\begin{enumerate}
\item \emph{Resemblance}: On updating the model from new data, any abrupt
model change can be difficult to interpret. For eg., suppose in the
previous model a predictor variable, \emph{age} and \emph{sex }are
significant and insignificant, respectively, but after an update in
current epoch, the significance of \emph{age }and \emph{sex }reversed.
Such abrupt changes can easily occur, due to multicollinearities and
often performed suboptimal search of objective space and, thus, many
possible solutions. However, interpreting adruptly changing solutions
is difficult and undesirable. 
\item \emph{Computation}: Since the data size will increase exponentially,
modeling all the data is computationally impractical. An effective
sequential modeling is required to break the computation into smaller
parts. 
\item \emph{Robust} (to noise and missing values): Some variables can be
unobserved (or scantily observed) in certain epochs. For example,
at Walmart several products get added/removed everyday or suspended
for some time. It is still essential to keep track of these dynamic
changes and appropriately handle the model.  Also, there can be
a large amount of noise occuring in an epoch (due to any external
event). For example, someone else using your Netflix for a short while,
often leading to your recommendations going of the chart. 
\item \emph{Accuracy}: Obtaining high prediction accuracy in a changing
and high-dimensional process is difficult. Specifically, overfitting
is a common issue in modeling a high-dimensional sparse model with
poor model selection, leading to a high test error. 
\end{enumerate}
It is, therefore, important to develop a new method that can sequentially
model a high-dimensional data streaming process, with inherent sparsity
and evolution. To that end, we propose a method called Inertial Regularization
and Selection (IRS). Similar to state-space models, IRS sequentially
models them over subsequent time epochs. A time epoch can be a day,
a week, or any other time interval at which the model is required
to be updated. 

IRS has an \emph{inertia }component in the objective loss function.
This inertia component resists abrupt model change, bringing \emph{resemblance}
with model learnt in earlier epochs. This also enables IRS to use
the prior knowledge learnt on the model so far, together with the
current epoch's data, for a yield an accurate model. It also keeps
the model smooth and robust to intermittent noisy observations. Besides,
unlike other state-space models, it works better in high-dimensional
and sparse problems due to its automatic optimal model selection property.
An optimal sparse model estimate prevents overfitting, and thus, lowers
test prediction errors.

IRS is a domain agnostic (generic) methodology that can handle any
large or small statistical learning problems with different specifications.
In this paper, it is developed for linear relationships and gaussian
errors, however, it can be easily extended to non-linear scenarios
using a link function. We show existing related methods as a special
case of IRS, viz. Kalman filters, (adaptive) Lasso and Bayesian models.
However, these models have certain critical limitations, especially
in terms of ability to handle high-dimensions, use of past information
(sequential modeling) and computation, discussed more in \S\ref{sec:Related-Work}
below, which is overcome by IRS. Through development of IRS, this
paper also shows the connection between Kalman Filters, regularization
methods, and Bayesian frameworks. As a by-product, we also show an
estimation procedure for Kalman filter which has significantly faster
computation in the usual case of data size being larger than the number
of predictors.

\medskip{}

In the remaining of this paper, we first review the related work in
this area in \S\ref{sec:Related-Work}. Thereafter, we show the
development of IRS in \S\ref{sec:Inertial-Regularization-and}.
This section provides the definition and elaboration of IRS, followed
by development of IRS regularization parameters (\S\ref{subsec:Regularization-parameters}).
Besides, within this section, equivalence of IRS with other methods
as special cases are shown in \S\ref{subsec:Lasso-equivalence}-\ref{subsec:Equivalent-Bayesian-Formulation}.
Further, we present a closed-form solution under orthogonality conditions
(in \S\ref{subsec:Closed-form-solution-under}) and an iterative
solution, for any general case, in Sec\@.~\ref{subsec:Iterative-estimation}.
An algorithm based on the iterative solution is postulated in \S\ref{subsec:Algorithm},
which also discusses various computational and implementation aspects
of IRS. Furthermore, we experimentally validate IRS's efficacy in
\S\ref{sec:Experimental-Validation}, and show its application
on a real online retail data in \S\ref{sec:Case-Study}. Finally,
we discuss the paper and conclude in \S\ref{sec:Discussion-and-Conclusion}.

\section{Related Work}
\label{sec:Related-Work}

Sequential modeling of statistical learning, specifically in context
of regression, was attempted by Gauss (1777\textendash 1885) more
than a century ago. However, his method, called as recursive least-squares
(RLS) estimation came into light several years later after a rediscovery
by \cite{plackett1950}. RLS works by estimating regression parameters
in a model with the available data and updates it when new data is
observed. However, it works if the model parameters are expected to
be stable, i.e. the model is not evolving. Moreover, since the entire
data until the present is used for the model, the effect of successive
new data on model starts to diminish. To account for this, a rolling
regression, which a data-pruning approach that keeps only a fixed
window of data for modeling. A gradual discounting of old data using
some decaying function is also used. Nevertheless, RLS still suffers
from its inability to effectively model an evolving process, and also
in handle high-dimensional data.

To better account for the model evolution, Kalman filters (KF) were
developed\cite{kalman1960}. KF is a sequential learning procedure,
where the model at any any epoch is estimated as a weighted average
of its predicted state from the previous state and the estimate from
the new data. For a consistent estimation, KF incorporates the model's
covariance for the model reestimations \textemdash{} the weights in
the weighted average proportional to this covariance \textemdash ,
but it is not designed to handle high-dimensional data. Other generalizations
of KF, viz. extended Kalman filter and unscented Kalman filter \cite{julier2004} were developed for nonlinear systems, but also
lacked in their ability to model high-dimensional data.

Ensemble Kalman filter (EnKF)\cite{evensen2003}, was developed for high-dimensional
situations. It is a combination of the sequential Monte Carlo and
Kalman filter. However, EnKFs are majorly focused on a proper approximation
of the model's covariance matrix in situations of high-dimension but
less data (via. Monte Carlo approximations). Here an ensemble a collection
of model realizations from different Monte Carlo generated data replications,
from which the model covariance is approximated as the sample covariance.
As can be noted, it does not address the issue of sparse model selection
under high-dimensionality. While some ``localization'' techniques
are some times used for dimension reduction\cite{ott2004},
for eg. in geophysics the model is decomposed into many overlapping
local patches based on their physical location, filtering is performed
on each local patch, and the local updates are combined to get the
complete model. However, this technique is not easy to generalize
to any model.

Particle filters (PF)\cite{gordon1993} are another class of method for
state-space modeling under nonlinearity. It is usually seen as an
alternative to KFs in nonlinear scenarios. But they severely suffer
from the ``curse of dimensionality'' due to its nonparametric nature
even for moderately large dimension\cite{liu2008}. \cite{lei2009}
developed an ensemble filtering method by combining the EnKF and PF
for high-dimensional settings. They used a similar ``localization''
technique described above, which limits its application.

State-space models, specifically Kalman filter, have been extensively
developed in the field of signal processing. To handle high-dimension
and sparsities in signals the methods extended the Kalman filters
to address them. \cite{vaswani2008} proposed a Kalman filtered compressed
sensing, which was further extended in \cite{carmi2010}, who directly
modified the Kalman filter equations using compressed sensing and
quasi-norm constraints to account for sparsity. However, these methods
lack theoretical support and has limited application. \cite{charles2011, charles2013} showed different signal regularization possibilities
for different signal dynamics scenarios. They also showed an extension
of \cite{candes2007} and \cite{candes2008} by adding an
$L_{2}$ penalty in the optimization objective. While this is close
to our objective, they used fixed penalty parameters, and did not
develop the methodology. Besides, \cite{charles2013} proprosed
a special bayesian formulation with Gamma prior distributions that
is good for the signal problems addressed in the paper but not generically
applicable. Moreover, similar to a fused lasso setup, a L1 penalty
to smoothen the dynamically changing system together with L1 regularization
for sparsity was also proposed by them and \cite{sejdinovic2010}. However,
these methods are difficult to optimize and, do not guarantee the
covergence from an erroneous initialization to a steady-state estimation
error.

In contrast to the existing methods, the proposed IRS can easily handle
linear and nonlinear (with linear link function) models in high-dimensional
space. As we will show in this paper, it provides an efficient sequential
modeling approach to accurate model such problems.

\section{Inertial Regularization and Selection (IRS)}
\label{sec:Inertial-Regularization-and}

\subsection{Definition and Notations\label{subsec:Definition-and-Notations}}

The following defines an evolutionary model for a data streaming process.

\begin{eqnarray}
\theta_{t} & = & F_{t}\theta_{t-1}+\nu_{t}\label{eq:model-eq-1}\\
y_{t} & = & X_{t}\theta_{t}+\epsilon_{t}\label{eq:model-eq-2}
\end{eqnarray}

Here $t$ denotes a time epoch for modeling. An epoch can be a day,
a week or any other time span, during which the model state is assumed
to not change. The observed data at $t$ is denoted by $X_{t}$ for
all independent variables (also referred to as predictors) and $y_{t}$
is the response. Eq.~\ref{eq:model-eq-2} is conventional statistical
model where the response is a linear function of predictors. $\theta_{t}$
is the model parameter and $\epsilon_{t}$ is an error term from a
normal distribution with mean 0 and covariance $W_{t}$, i.e. $\epsilon_{t}\sim N(0,W_{t})$.
Besides, $\theta_{t}$ also specifies the model state at epoch $t$.

While Eq.~\ref{eq:model-eq-2} models the response, Eq.~\ref{eq:model-eq-1}
captures the any change in the state. Often these state changes are
small and/or smooth (gradual). As also mentioned before, these changes
are called as evolution. Eq.~\ref{eq:model-eq-1} also has a linear
structure, where $F_{t}$ is called state transition parameter and
$\nu_{t}$ is the state change process noise, assumed to be $\sim N(0,Q_{t})$.
Besides, in the rest of the paper we refer Eq.~\ref{eq:model-eq-1}
as state-model and Eq.~\ref{eq:model-eq-2} as response-model. 

The data sample size and the number of predictor variables at $t$
is denoted as $n_{t}$ and $p_{t}$, respectively. Thus, $X_{t}\in\mathbb{R}^{n_{t}\times p_{t}}$
and $y_{t}\in\mathbb{R}^{n_{t}}$. Additionally, the covariance of
$\theta_{t}$ is denoted as $\Sigma_{t}$. 

\subsection{IRS Model\label{subsec:IRS-Model}}

The objective of IRS model is to sequentially learn a statistical
model from a streaming data from a high dimensional process. For that,
the following objective function is proposed that, upon minimization,
yields an optimal model estimate at any epoch $t$.

\begin{multline}
\mathcal{L}(\theta_{t})=\frac{1}{2n_{t}}(y_{t}-X_{t}\theta_{t})^{T}W_{t}^{-1}(y_{t}-X_{t}\theta_{t})+\\
\frac{\tau}{2p_{t}}(\theta_{t}-F_{t}\theta_{t-1})^{T}\Sigma_{t|t-1}^{-1}(\theta_{t}-F_{t}\theta_{t-1})+\frac{\lambda}{p_{t}}\sum_{i=1}^{p_{t}}\frac{|\theta_{t_{i}}|}{|\theta_{t_{i}}^{*}|}\label{eq:main-loss-main}
\end{multline}

The above objective comprises of three components: 

\medskip{}

a) The first term is a sum of residual squares scaled by the residual
error covariance, 

\medskip{}

b) The second term is a $L_{2}$-norm regularization that penalizes
any drastic change in the model estimate. A change here implies the
difference in the model state at $t$ from the expected model state
at $t$, given the state at $t-1$ $(=F_{t}\theta_{t-1})$. The second
term is, thus, called an inertia component because it resists drastic
changes, and in effect, keeps the updated model estimate close to
the anticipated model derived from the past model. Besides, this inertia
also resists a change due to any noisy observations. These properties
are a result of using $\Sigma_{t|t-1}^{-1}$ as an adaptive penalty
for an effective model updation, and also helps the inertia component
differentiate an abnormal observation from noise (discussed in \S\ref{subsec:Regularization-parameters}
and \ref{sec:Discussion-and-Conclusion}).

\medskip{}

c) The third term is a $L_{1}$-norm penalty term for an optimal model
selection. Although, the $L_{1}$ penalty is imposed ``spatially'',
i.e. only on the model parameters at epoch $t$, the selected model
will still resemble with the selected model in the past. This is in
part due to the inertial component and an adaptive penalty parameter
$\theta_{t}^{*}$. This aspect is further elaborated in \S\ref{subsec:Regularization-parameters}.

\medskip{}

In Eq.~\ref{eq:main-loss-main}, $\theta_{t-1}$ is unknown, but
it is shown in Appendix-A that it suffices to use the previous epoch's
model estimate instead of their true values. This is a founding property
for IRS, that allows a sequential (online) modeling. $\theta_{t-1}$
can be, thus, replaced with its estimate $\hat{\theta}_{t-1}$. For
a succinct notation, we denote $\hat{\theta}_{t|t-1}=F_{t}\hat{\theta}_{t-1}$,
where $\hat{\theta}_{t|t-1}$ is the estimate model state at $t$
given $\theta_{t-1}$. Similarly, the expected covariance of $\hat{\theta}_{t|t-1}$,
denoted by $\hat{\Sigma}_{t|t-1}$, is derived from the previous state's
covariance, where $\hat{\Sigma}_{t|t-1}=F_{t}\hat{\Sigma}_{t-1}F_{t}^{T}+Q_{t}$.
Likewise, $\theta_{t}^{*}$ is replaced by $\hat{\theta}_{t}^{*}$.
The resultant online objective function for sequential modeling is,
thus,

\begin{multline}
\hat{\theta}_{t}=\arg\min_{\theta_{t}}\frac{1}{2n_{t}}(y_{t}-X_{t}\theta_{t})^{T}W_{t}^{-1}(y_{t}-X_{t}\theta_{t})+\\
\frac{\tau}{2p_{t}}(\theta_{t}-\hat{\theta}_{t|t-1})^{T}\hat{\Sigma}_{t|t-1}^{-1}(\theta_{t}-\hat{\theta}_{t|t-1})+\frac{\lambda}{p_{t}}\sum_{i=1}^{p_{t}}\frac{|\theta_{t_{i}}|}{|\hat{\theta}_{t_{i}}^{*}|}\label{eq:main-loss-main-3}
\end{multline}

Besides, hyperparameters $\lambda$ and $\tau$ regulates the amount
of penalty imposed for limiting the model size (sparsity) and closeness
to the expected model given the past, respectively. $\lambda$ and
$\tau$ are later shown to be independent of epoch $t$ unless there
is any true drastic change in the system. This is an important property
because it removes the computational overhead of searching optimal
$\lambda$ and $\tau$for every epoch. Besides, each term in the objective
is normalized by their size, $n_{t}$ or $p_{t}$, to meet this end. 

\begin{figure}
\begin{centering}
\begin{minipage}[t]{0.45\columnwidth}%
\begin{center}
\subfloat[Elastic-net\label{fig:enet-illustration}]{\begin{centering}
\includegraphics[scale=0.3]{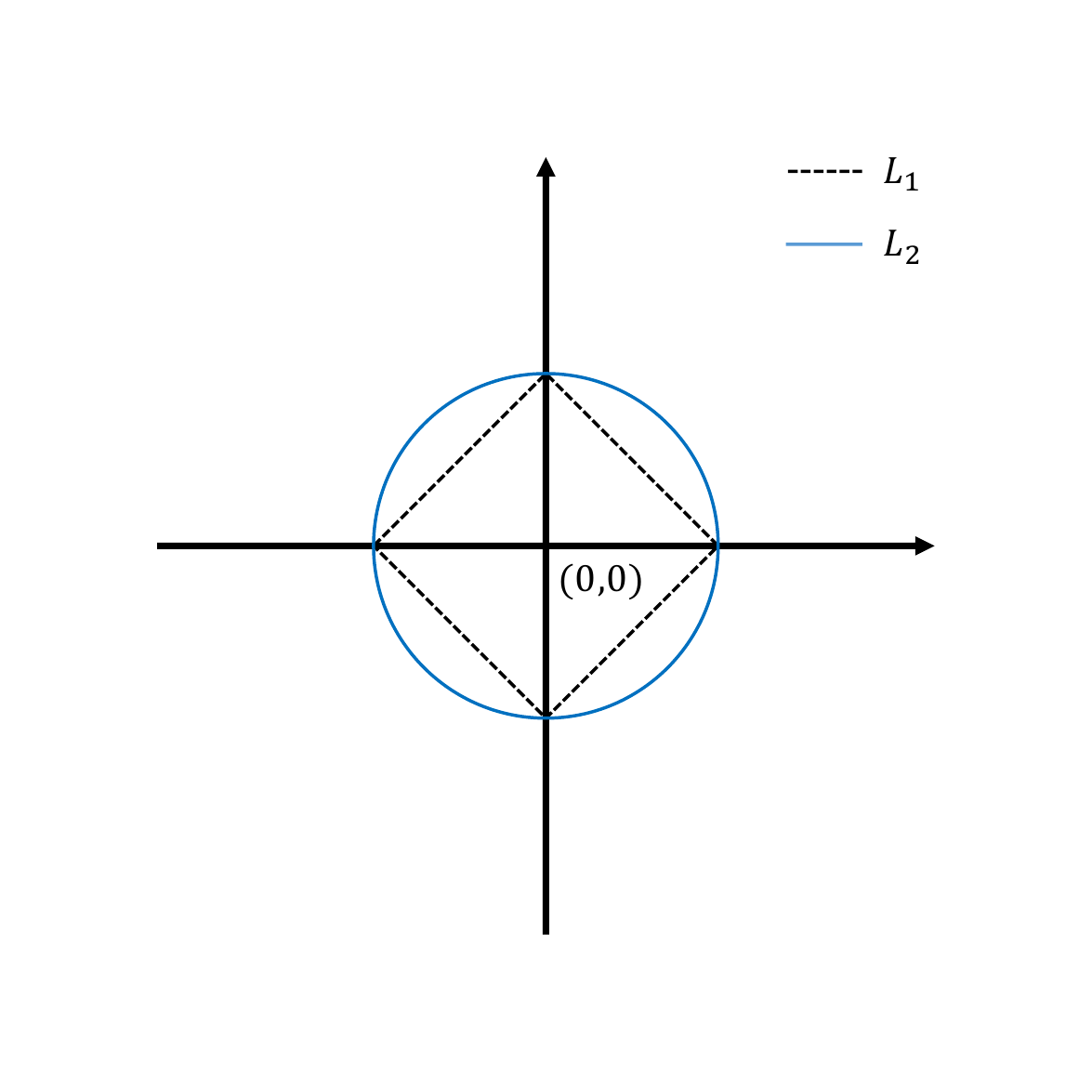}
\par\end{centering}
}
\par\end{center}%
\end{minipage}\hspace{20bp}%
\begin{minipage}[t]{0.45\columnwidth}%
\begin{center}
\subfloat[Inertial Regularization and Selection (IRS) \label{fig:ilasso-illustration}]{\begin{centering}
\includegraphics[scale=0.3]{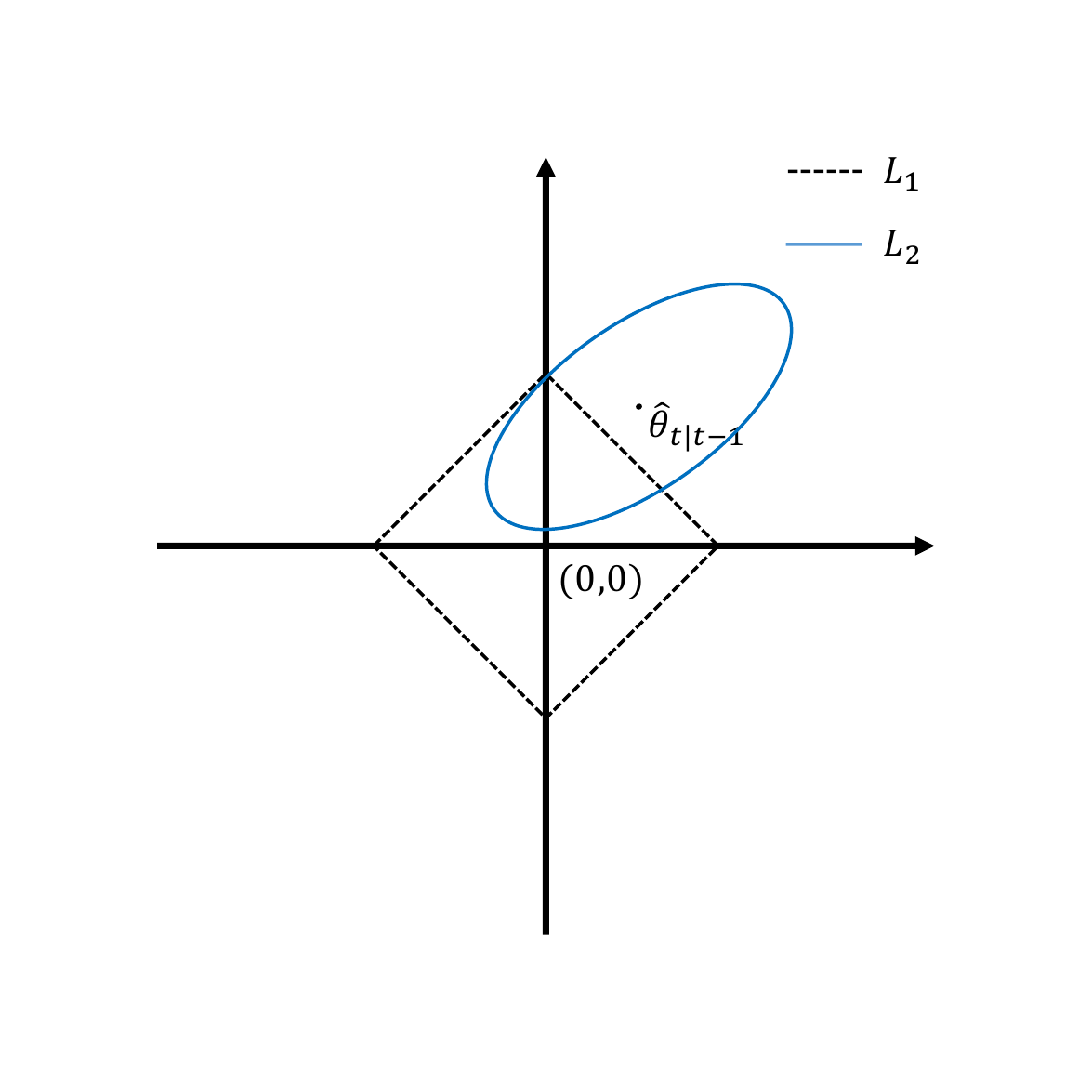}
\par\end{centering}
}
\par\end{center}%
\end{minipage}
\par\end{centering}
\caption{Graphical illustration of proposed Inertial Regularization and Selection
(IRS), and contrasting it with Elastic-net.}
\end{figure}

On another note, Eq.~\ref{eq:main-loss-main-3} bears close resemblance
with an Elastic-net formulation (Zou and Hastie, 2005). Their difference,
illustrated graphically in Fig.~\ref{fig:enet-illustration}-\ref{fig:ilasso-illustration}
for a simple 2-dimensional model, further explains and highlights
the differentiating properties of IRS. As shown in the figures, the
$L_{2}$ penalty term in Elastic-net is centered around 0, while IRS's
$L_{2}$ regularization is around, $\hat{\theta}_{t|t-1}$, the model
estimate given the past or the prior knowledge. Moreover, IRS attempts
to find a sparse model that also matches the prior knowledge. This
reinstates the previous claim that despite of a spatial $L_{1}$ penalty
the resultant model resembles the model derived from the past.

\subsection{Why $\Sigma_{t|t-1}$ and $\theta_{t}^{*}$ adapted regularization?\label{subsec:Regularization-parameters}}

In a sequential regularization model, it is critical to update the
regularization penalties in a data driven manner. Otherwise, a repetitive
regularization optimization for each epoch can be computationally
cumbersome. We, therefore, use estimates of $\Sigma_{t|t-1}$ and
$\theta_{t}^{*}$, which are derived from the data and have the necessary
properties, discussed below, that results into an accurate model estimation.

For the inertia part, we want close to the expected value of $\theta_{t}$
given $\hat{\theta}_{t-1}$, i.e. $\hat{\theta}_{t|t-1}$. But the
extent of regularization should be different for different $\theta_{t_{i}}$'s.
The $\theta_{t_{i}}$ with higher expected variance should be regularized
lesser. This is because, higher variance indicates that the $\theta_{t_{i}}$'s
estimate from the past is not reliable. Thus, it should be learnt
more from a recent data from epoch $t$ than keeping close to the
past. And it is vice-versa if the expected variance of $\theta_{t_{i}}$
is small \textemdash{} tend to keep it close to a reliable past knowledge.
Therefore, the inertia should be weighted by the inverse of the variance.
Besides, as the correlations between the model parameters should be
taken into account, as well, to adjust the changes in correlated parameters,
we use $\Sigma_{t|t-1}^{-1}$ as an adaptive inertia weight.

In addition to the above intuition, $\Sigma_{t|t-1}^{-1}$ also stabilizes
the inertial component. The inertial component can be regarded as
a sum of residual squares from Eq.~\ref{eq:model-eq-1}, i.e. we
aim to minimize the residuals of $(\theta_{t}-\hat{\theta}_{t|t-1})$.
Thus, the residual sum of squares should be scaled by its covariance,
where $\Sigma_{t|t-1}=\text{cov}(\theta_{t}-\hat{\theta}_{t|t-1})$. 

Besides, an adaptive penalty is applied on the $L_{1}$ norm for model
selection. As suggested in \cite{zou2006}, inverse of a reasonable estimate
for $\theta$ provides a good adaptive penalty term for $L_{1}$ norm.
\cite{zou2006} suggested an OLS estimate for a Lasso problem as a good
estimate for the adaptive penalty, as it brings estimation consistency
and guarantees oracle properties. As shown later, in \S\ref{subsec:Lasso-equivalence},
the IRS objective function in Eq.~\textbf{\ref{eq:main-loss-main-3}
}is equivalent to a Lasso problem. From this Lasso equivalence, it
can be seen that the OLS estimate desired for adaptively penalizing
$\theta_{t}$ is same as an only inertia regularized estimate for
$\theta_{t}$ , denoted as $\hat{\theta}_{t}^{*}$. The expression
for $\hat{\theta}_{t}^{*}$ is derived and given in Appendix-B. We,
therefore, use $\lambda/|\hat{\theta}_{t_{i}}^{*}|$ as penalty for
the $\theta_{t_{i}}$'s $L_{1}$ regularization. 

\subsection{Lasso equivalence\label{subsec:Lasso-equivalence}}

Here we show a Lasso equivalent formulation of IRS. This is an important
property because it shows IRS can inherit from a well developed pool
of Lasso based techniques. In fact, the use of $\theta_{t}^{*}$ as
the adaptive $L_{1}$ penalty for model selection is drawn from this
Lasso equivalence. Besides, the estimation algorithm for IRS developed
in \S\ref{subsec:Solution} is also inspired from Lasso solutions.

The following shows the Lasso equivalence using a transformation and
augmentation of data $X_{t}$ and $y_{t}$.

\begin{equation}
\begin{split}\tilde{X}_{t}=\left[\begin{array}{c}
\frac{1}{\sqrt{2n_{t}}}W_{t}^{-\frac{1}{2}}X_{t}\\
\sqrt{\frac{\tau}{2p_{t}}}\Sigma_{t|t-1}^{-\frac{1}{2}}
\end{array}\right];\:\: & \tilde{y}_{t}=\left[\begin{array}{c}
\frac{1}{\sqrt{2n_{t}}}W_{t}^{-\frac{1}{2}}y_{t}\\
\sqrt{\frac{\tau}{2p_{t}}}\Sigma_{t|t-1}^{-1/2}\hat{\theta}_{t|t-1}
\end{array}\right]\end{split}
\label{eq:Xy-augmentation}
\end{equation}

Using transformed data in Eq.~\ref{eq:Xy-augmentation}, we can rewrite
the objective function in Eq.~\ref{eq:main-loss-main-3} as a Lasso
problem, as given below (see Appendix-C for proof).

\begin{equation}
\mathcal{L}_{\hat{\theta}_{t-1}}(\theta_{t})=\left(\tilde{y}_{t}-\tilde{X}_{t}\theta_{t}\right){}^{T}\left(\tilde{y}_{t}-\tilde{X}_{t}\theta_{t}\right)+\frac{\lambda}{p_{t}}\sum_{i=1}^{p_{t}}\frac{|\theta_{t_{i}}|}{|\hat{\theta}_{t_{i}}^{*}|}\label{eq:Lasso-equi}
\end{equation}

\subsection{Kalman Filter as a special case\label{subsec:Kalman-Filter-equivalence}}

The Kalman Filter is found to be a special case of IRS. This finding
highlights the commonality between filtering methods and regularization.

When $\lambda=0$, IRS becomes equivalent to a more general Kalman
filter which has a weight adjusted state update, i.e. a weighted average
of the state estimate from past and a Kalman estimate from the current
data. The following shows the IRS estimate expression as equal to
a Kalman filter expression (see Appendix-D for the proof).

\begin{eqnarray}
\hat{\theta}_{t}^{*} & = & \left[X_{t}^{T}W_{t}^{-1}X_{t}+\tau^{*}\Sigma_{t|t-1}^{-1}\right]^{-1}\cdot\nonumber \\
 &  & \left[X_{t}^{T}W_{t}^{-1}y_{t}+\tau^{*}\Sigma_{t|t-1}^{-1}\hat{\theta}_{t|t-1}\right]\nonumber \\
 & = & \hat{\theta}_{t|t-1}+\frac{1}{\tau^{*}}K_{t}(y_{t}-X_{t}\hat{\theta}_{t|t-1})\label{eq:kalman-1}
\end{eqnarray}

where, $K_{t}=\Sigma_{t|t-1}X_{t}^{T}\left(W_{t}+\frac{1}{\tau^{*}}X_{t}\Sigma_{t|t-1}X_{t}^{T}\right)^{-1}$
, also known as Kalman gain, and $\tau^{*}=\tau n_{t}/p_{t}$. As
we can see above (Eq.~\ref{eq:kalman-1}), the expression for $\theta_{t}$
estimate is same as a conventional (unweighted) Kalman filter when
$\tau^{*}=1$, and otherwise a weighted Kalman filter when $\tau^{*}$
is any other positive value. Besides, when $\tau^{*}=1$, the covariance
of $\hat{\theta}_{t}^{*}$ can be shown to be same as in Kalman filter.

\begin{eqnarray}
\Sigma_{t}^{*} & = & \text{cov}\left\{ \left[X_{t}^{T}W_{t}^{-1}X_{t}+\Sigma_{t|t-1}^{-1}\right]^{-1}\cdot\right.\nonumber \\
 &  & \left.\left[X_{t}^{T}W_{t}^{-1}y_{t}+\Sigma_{t|t-1}^{-1}\hat{\theta}_{t|t-1}\right]\right\} \nonumber \\
 & = & (I-K_{t}X_{t})\Sigma_{t|t-1}\label{eq:kalman-2}
\end{eqnarray}

It can be noted that, for estimating $\theta_{t}$ and $\Sigma_{t}$,
the Kalman filter expression requires inverse of $K_{t}$. $K_{t}$,
however, is a $n_{t}\times n_{t}$ matrix, inverting which can be
computationally expensive when data size, $n_{t}$, is large. We provide
an implementation for Kalman filter in Appendix-D, which is significantly
faster for a usual scenario of $p_{t}<n_{t}$.

\subsection{Equivalent Bayesian Formulation\label{subsec:Equivalent-Bayesian-Formulation}}

The Bayesian formulation of IRS will have the following prior for
$\theta$,
\begin{multline}
p_{\lambda,\tau}(\theta_{t})\propto\exp-\left(\frac{\lambda}{p_{t}}\sum_{i=1}^{p_{t}}\frac{|\theta_{t_{i}}|}{|\hat{\theta}_{t_{i}}^{*}|}\right)\cdot\\
\exp\left(-\frac{\tau}{2p_{t}}(\theta_{t}-\hat{\theta}_{t|t-1})^{T}\Sigma_{t|t-1}(\theta_{t}-\hat{\theta}_{t|t-1})\right)\label{eq:bayesian-prior}
\end{multline}

This prior is a joint distribution of a double-exponential distribution
for the $L_{1}$ penalty and a Gaussian centered around $\hat{\theta}_{t|t-1}$
and scaled with the covariance $\Sigma_{t|t-1}$ for the inertial
$L_{2}$ penalty. The parameters $\theta_{t}$ can be estimated as
the Bayes posterior mode (see Appendix-A).

\subsection{Model covariance\label{subsec:Model-covariance}}

Due to an $L_{1}$ norm term in the loss function, it is difficult
to obtain an accurate estimate for the model covariance, $\Sigma_{t}$.
Bootstraping can be a numerical way to estimate it. However, bootstrapping
can be time consuming. Therefore, we propose an approximate closed
form estimate, similar to the one proposed in \cite{tibshirani1996}. In
\cite{tibshirani1996}, the $L_{1}$ penalty $\sum|\theta_{i}|$ was replaced
with $\sum\theta_{i}^{2}/|\theta_{i}|$, and the covariance matrix
was approximated at the lasso estimate, $\hat{\theta}$, as a ridge
regression expression. 

This estimation, however, gives a 0 variance estimate if $\hat{\theta}_{i}$'s
equal to 0. This is not desired in IRS, because a variance estimate
for each $\theta_{i}$ is necessary for the inertial component. We,
therefore, provide a slightly modified covariance expression, where
$|\theta_{t_{i}}|$ is replaced by $\theta_{t_{i}}^{2}/|\hat{\theta}_{t_{i}}^{*}|$.
Thus, an approximate covariance estimate is given as,

\begin{eqnarray}
\Sigma_{t} & = & A_{t}^{-1}\left(X_{t}^{T}W_{t}^{-1}X_{t}+\tau^{*2}\Sigma_{t|t-1}^{-1}\right)A_{t}^{-1}\label{eq:Sigma_t_exp}
\end{eqnarray}
where,
\begin{eqnarray}
A_{t} & = & X_{t}^{T}W_{t}^{-1}X_{t}+\lambda D_{t}^{-1}+\tau^{*}\Sigma_{t|t-1}^{-1}\\
D_{t} & = & \text{diag}\left(I(\hat{\theta}_{t_{i}}\neq0)|\hat{\theta}_{t_{i}}||\hat{\theta}_{t_{i}}^{*}|+\right.\nonumber \\
 &  & \left.(1-I(\hat{\theta}_{t_{i}}\neq0))|\hat{\theta}_{t_{i}}^{*}|^{2};\,i=1,\ldots,p_{t}\right)
\end{eqnarray}
 and $\tau^{*}=\tau n_{t}/p_{t}$.

\subsection{Model Estimation\label{subsec:Solution}}

In this section, we first develop a closed form solution under orthogonality
conditions, and then present an iterative proximal gradient estimation
method for any general case. An algorithm is then laid out for implementing
IRS in \S\ref{subsec:Algorithm}, where the various computational
and implementation aspects are discussed.

\subsubsection{Closed-form solution under orthogonality\label{subsec:Closed-form-solution-under}}

The developed estimation technique possesses a closed-form solution
if, (a) the predictors are orthonormal ($X_{t}^{T}X_{t}=I$), i.e.
all columns in $X_{t}$ are linearly independent of each other, (b)
the error of response model is i.i.d., $W_{t}=w_{t}^{2}I$, and, (c)
the error covariance of the state model is i.i.d., $Q_{t}=q_{t}^{2}I$,
which is often implied from if (a) is true.

Orthogonality conditions, a) $X_{t}^{T}X_{t}=I$, b) $W_{t}=w_{t}^{2}I$,
and c) $Q_{t}=q_{t}^{2}I$. Under these conditions, $\Sigma_{t}$,
and in turn $\Sigma_{t|t-1}$, will be diagonal matrices. Suppose
we denote, $\Sigma_{t|t-1}=\boldsymbol{\rho}_{t}I$, where $\rho_{t_{i}}$
is the expected variance of $\theta_{t_{i}}$.

Plugging these conditions into Eq.~\ref{eq:OLS-lasso}, we get the
OLS estimate expression as,

\begin{eqnarray}
\hat{\theta}_{t}^{*} & = & \left(\frac{1}{w_{t}^{2}}I+\tau^{*}(\boldsymbol{\rho}_{t}I)^{-1}\right)^{-1}\left(\frac{1}{w_{t}^{2}}X_{t}^{T}y_{t}+\tau^{*}(\boldsymbol{\rho}_{t}I)^{-1}\hat{\theta}_{t|t-1}\right)\nonumber \\
 & = & (\boldsymbol{\rho}_{t}^{*}I)\left(\frac{1}{w_{t}^{2}}X_{t}^{T}y_{t}+\tau^{*}(\boldsymbol{\rho}_{t}I)^{-1}\hat{\theta}_{t|t-1}\right)\label{eq:OLS-Lasso-Ortho}
\end{eqnarray}

where, $\tau^{*}=\frac{\tau n_{t}}{p_{t}}$ and $\frac{1}{\rho_{t_{i}}^{*}}=\frac{1}{w_{t}^{2}}+\frac{\tau^{*}}{\rho_{t_{i}}},i=1,\ldots,p_{t}$.

We will use this expression to derive the IRS closed-form solution
under orthogonality. From the Lasso equivalent formulation of IRS,
the parameter estimate can be computed by minimizing, Eq.~\ref{eq:Lasso-equi},
i.e.,

\begin{equation}
\mathcal{L}(\theta_{t})=\left(\tilde{y}_{t}-\tilde{X}_{t}\theta_{t}\right){}^{T}\left(\tilde{y}_{t}-\tilde{X}_{t}\theta_{t}\right)+\frac{\lambda}{p_{t}}\sum_{i=1}^{p_{t}}\frac{|\theta_{t_{i}}|}{|\hat{\theta}_{t_{i}}^{*}|}\label{eq:Lasso-equi-1}
\end{equation}

\begin{eqnarray}
\arg\min_{\theta_{t}} &  & -2\tilde{y}_{t}^{T}\tilde{X}_{t}\theta_{t}+\theta_{t}^{T}\tilde{X}_{t}^{T}\tilde{X}_{t}\theta_{t}+\frac{\lambda}{p_{t}}\sum_{i=1}^{p_{t}}\frac{|\theta_{t_{i}}|}{|\hat{\theta}_{t_{i}}^{*}|}\nonumber \\
= &  & -2\left(\frac{1}{2n_{t}}y_{t}^{T}W_{t}^{-1}X_{t}\theta_{t}+\frac{\tau}{2p_{t}}\hat{\theta}_{t|t-1}^{T}\Sigma_{t|t-1}^{-1}\theta_{t}\right)+\nonumber \\
 &  & \frac{1}{2n_{t}}\theta_{t}^{T}X_{t}^{T}W_{t}^{-1}X_{t}\theta_{t}+\frac{\tau}{2p_{t}}\theta_{t}^{T}\Sigma_{t|t-1}^{-1}\theta_{t}+\frac{\lambda}{p_{t}}\sum_{i=1}^{p_{t}}\frac{|\theta_{t_{i}}|}{|\hat{\theta}_{t_{i}}^{*}|}\nonumber \\
\propto &  & -\left(\frac{1}{w_{t}^{2}}y_{t}^{T}X_{t}\theta_{t}+\tau^{*}(\boldsymbol{\rho}_{t}I)^{-1}\hat{\theta}_{t|t-1}^{T}\theta_{t}\right)+\nonumber \\
 &  & \frac{1}{2w_{t}^{2}}\theta_{t}^{T}\theta_{t}+\tau^{*}(\boldsymbol{\rho}_{t}I)^{-1}\theta_{t}^{T}\theta_{t}+\frac{\lambda n_{t}}{p_{t}}\sum_{i=1}^{p_{t}}\frac{|\theta_{t_{i}}|}{|\hat{\theta}_{t_{i}}^{*}|}\nonumber \\
= &  & -\left(\frac{1}{w_{t}^{2}}y_{t}^{T}X_{t}+\tau^{*}(\boldsymbol{\rho}_{t}I)^{-1}\hat{\theta}_{t|t-1}^{T}\right)\theta_{t}+\nonumber \\
 &  & \frac{1}{2}\left(\frac{1}{w_{t}^{2}}+\tau^{*}(\boldsymbol{\rho}_{t}I)^{-1}\right)\theta_{t}^{T}\theta_{t}+\frac{\lambda n_{t}}{p_{t}}\sum_{i=1}^{p_{t}}\frac{|\theta_{t_{i}}|}{|\hat{\theta}_{t_{i}}^{*}|}\nonumber \\
= &  & -(\boldsymbol{\rho}_{t}^{*}I)^{-1}\hat{\theta}_{t}^{*T}\theta_{t}+\frac{1}{2}(\boldsymbol{\rho}_{t}^{*}I)^{-1}\theta_{t}^{T}\theta_{t}+\frac{\lambda n_{t}}{p_{t}}\sum_{i=1}^{p_{t}}\frac{|\theta_{t_{i}}|}{|\hat{\theta}_{t_{i}}^{*}|}\nonumber \\
= &  & \sum_{i=1}^{p_{t}}-\frac{1}{\rho_{t_{i}}^{*}}\hat{\theta}_{t_{i}}^{*}\theta_{t_{i}}+\frac{1}{2\rho_{t_{i}}^{*}}\theta_{t_{i}}^{2}+\frac{\lambda n_{t}}{p_{t}}\frac{|\theta_{t_{i}}|}{|\hat{\theta}_{t_{i}}^{*}|}\label{eq:Ortho-Lasso-obj}
\end{eqnarray}
\textbackslash{} 

On solving this for both cases when, $\theta_{i}^{(t)}\ge0$ and $\leq0$,
we get the following solution,

\begin{eqnarray}
\hat{\theta}_{t_{i}} & = & \text{sgn}(\hat{\theta}_{t_{i}}^{*})\left(\left|\hat{\theta}_{t_{i}}^{*}\right|-\frac{\lambda n_{t}}{p_{t}}\frac{\rho_{t_{i}}^{*}}{|\hat{\theta}_{t_{i}}^{*}|}\right)^{+}\label{eq:Closed-form-solution}
\end{eqnarray}

where, $(\cdot)^{+}$ is a soft-thresholding function. 

The components in the solution expression (Eq.~\ref{eq:Closed-form-solution})
can be interpreted as a shrinkage on the inertia regularized parameter
(OLS) estimate, $\hat{\theta}_{t_{i}}^{*}$, such that the amount
of shrinkage is determined by, a) the magnitude of $\hat{\theta}_{t_{i}}^{*}$,
for an adaptive variable selection from $L_{1}$ penalty, and b) $\rho_{t_{i}}^{*}$,
which is proportional to the expected variance of the OLS estimate,
$\hat{\theta}_{t_{i}}^{*}$, thus, IRS strongly shrinks $\hat{\theta}_{t_{i}}^{*}$
if its expected variance is high. 

Therefore, parameters with high variance and low magnitude will be
removed. The addition of variance element boosts the model selection
capability to select a better model, resulting into a lower prediction
errors.

\subsubsection{Iterative estimation\label{subsec:Iterative-estimation}}

Since, most commonly found problems do not satisfy an orthogonality
condition, we develop a proximal gradient method to estimate the parameters
in a general case. 

\begin{eqnarray*}
\mathcal{L}(\theta_{t}) & = & \left\{ \frac{1}{2n_{t}}\left(y_{t}-X_{t}\theta_{t}\right){}^{T}W_{t}^{-1}\left(y_{t}-X_{t}\theta_{t}\right)\right.+\\
 &  & \underbrace{\left.\frac{\tau}{2p_{t}}(\theta_{t}-\hat{\theta}_{t|t-1})^{T}\Sigma_{t|t-1}^{-1}(\theta_{t}-\hat{\theta}_{t|t-1})\right\} }_{f(\theta_{t})\rightarrow\text{differentiable}}\\
 &  & +\frac{\lambda}{p_{t}}\sum_{i=1}^{p_{t}}\frac{|\theta_{t_{i}}|}{|\hat{\theta}_{t_{i}}^{*}|}_{g(\theta_{t})\rightarrow\text{non-differentiable}}\\
 & = & f(\theta_{t})+g(\theta_{t})
\end{eqnarray*}

Since, $g$ is a $L_{1}$ function, an iterative proximal gradient
algorithm will perform the following update, 

\begin{eqnarray}
\theta_{t_{i}}^{l+1} & \leftarrow & \text{Prox}_{\nu}\left(\theta_{t_{i}}^{l}-s^{l}\nabla f(\theta_{t_{i}}^{l})\right)\label{eq:Prox-update-base}
\end{eqnarray}

where, $\nu=\frac{\lambda s^{l}}{p_{t}|\hat{\theta}_{t_{i}}^{*}|}$
and $\text{Prox}_{\nu}(x)=\begin{cases}
x-\nu & ;x\geq\nu\\
0 & ;|x|<\nu\\
x+\nu & ;x\leq-\nu
\end{cases}$ , $s^{l}$ is a step size for iteration $l$ (the step size can be
kept same for all $l$ or decreased with increasing $l$) and 
\begin{equation}
\nabla f(\theta_{t}^{l})=-\frac{1}{n_{t}}(X_{t})^{T}W_{t}^{-1}(y_{t}-X_{t}\theta_{t}^{l})+\frac{\tau}{p_{t}}\Sigma_{t|t-1}^{-1}(\theta_{t}^{l}-\hat{\theta}_{t|t-1})\label{eq:Prox-gradient}
\end{equation}

Note that, any other estimation technique, for example, subgradient
method or Least-Angle Regression (LARS), for estimating $\theta$.
We chose proximal gradient method because of its ease of understanding
and implementation. Besides, it performs well in most circumstances.

\subsubsection{Algorithm\label{subsec:Algorithm}}

In this section, we present Algorithm-\ref{alg:IRS-algo} for implementing
the iterative estimation method discussed above. The algorithm takes
in the predictor matrices, $X_{t}$, and the responses, $y_{t}$,
as they are collected for epoch $t=1,2,\ldots$. The $X_{t}$'s are
standardized within each epoch $t$, such that the columns in $X_{t}$
has a mean of 0 and variance of 1. $y_{t}$, on the other hand, is
centered to remove an intercept from the model. 

The algorithm assumes $F_{t}$ and $Q_{t}$ as known inputs. In practice,
$F_{t}$ is often set as equal to an identity matrix, $F_{t}=I$,
and $Q_{t}$ as $\acute{o}^{2}I$, where $\acute{o}$ is a small positive
number. Besides, the response error, $\epsilon_{t}$, in Eq.~\ref{eq:model-eq-2},
is assumed as i.i.d., i.e. $W_{t}=w^{2}I$, and $p_{t}=p,\forall t$.
While these are reasonable assumptions for most applications, it can
be easily extended for circumstances outside of these assumptions.

As shown in Algorithm-\ref{alg:IRS-algo}, the IRS estimation process
starts from line-24. To initialize, an OLS estimate is computed, and
residual sum of squares is used for obtaining $w^{2}$. Besides, the
covariance is initialized as identity. Thereafter, for any $t$, $\hat{\theta}_{t|t-1}$,
$\Sigma_{t|t-1}$ and $w^{2}$ (for approximating $W_{t}$) are computed
and passed into \textsc{ProximalDescent} function, alongwith the
data and regularization parameters, $(\lambda,\tau)$. This function
implements the proximal gradient steps shown above in \S\ref{subsec:Iterative-estimation}.
Note that, it takes inverse of $\Sigma_{t|t-1}$ to save computation.
Besides, it is recommended to have relatively larger data size in
the initial epoch to avoid yielding a sub-optimal initial model.

A common issue in regularization models is tuning the penalty parameters,
viz. $(\lambda,\tau)$ in IRS. It would have been even more challenging
if IRS required to (re)tune $(\lambda,\tau)$ for each estimation
epoch. However, due to the adaptive regularization developed in \S\ref{subsec:Regularization-parameters},
it can be shown that the expected value of the loss function, $\mathcal{L}_{\hat{\theta}_{t-1}}(\theta_{t})$,
remains a constant, i.e. independent of epoch t (see Appendix \textbf{??}).
This implies that an appropriately chosen $(\lambda,\tau)$ can be
used for any $t=1,2,\ldots$.

Therefore, for IRS, we can use a \emph{k}-fold cross validation (CV)
on a two-dimensional grid search for $(\lambda,\tau)$, and model
first few epochs for each grid point and measure their prediction
errors. We used a 10-fold CV on first three epochs for $(\lambda,\tau)$
selection. Zou and Hastie (2005) have described this approach in detail
and provided other possible selection approaches. 

Computational time is another common issue in estimation procedures.
The developed proximal gradient based algorithm has a fast convergence,
which can be further boosted by changing the stepsize, $s$, for each
iteration in line 11-18. This is common practice, with one simple
approach as starting with a large value for $s$ and gradually decreasing
it after each iteration. Another approach is to adaptively increase
or decrease $s$ based on an increase or decrease in the residual
error from $\theta_{t}^{l}$. Using this, the proximal gradient was
found to converge in less than 50 iterations for upto 1000-dimensional
models. 

Besides, in implementation, operations in lines 13-16 can be vectorized
and approximate matrix inversion methods can be used for a faster
computation. Furthermore, accelerated proximal gradient and stochastic
proximal gradient methods can be used to boost the computation.

\begin{algorithm}
\begin{algorithmic}[1] 
	\INPUT $X_t,y_t,\hat{\theta}_{t-1},\Sigma_{t-1}, \lambda, \tau, Q_t; t=1,2,\ldots$
	
	\algrule
	\Function{Prox}{$x,\nu$}
		\State{$x \gets \begin{cases}x-\nu & x>\nu\\0 & |x|\leq\nu\\x+\nu & x<-\nu\end{cases}$}
	\State \Return {$x$}
	\EndFunction

	\Statex
	
	\Function{ProximalDescent}{$X_t, y_t, \hat{\theta}_{t|t-1},\newline \Sigma_{t|t-1}^{-1}, w^2, \lambda, \tau$}
		\State{$n_t \gets \arg_{n_{t}} X_{t}\in\mathbb{R}^{n_{t}\times p}$}
		\State{$\tau^* \gets \tau n_t/p$}
		\State{$W_t \gets w^2 I_{n_t}$}
		\State{$\hat{\theta}_{t}^* \gets (X_t^TW_t^{-1}X_t + \tau^*\Sigma^{-1}_{t|t-1})^{-1}(X_t^TW_t^{-1}{y}_t + \tau^*\Sigma^{-1}_{t|t-1}\hat{\theta}_{t|t-1})$}
		\Statex
		\Initialize{$l=0, \theta_t^0 = \hat{\theta}_{t}^*, s$}
		\While {$l < L$}
			\State{$\nabla f(\theta_t^l) \gets \frac{-1}{n_t}X^T_t W_t^{-1} (Y_t-X_t\theta_t^l) + \frac{\tau}{p}\Sigma_{t|t-1}^{-1}(\theta_t^l - \hat{\theta}_{t|t-1})$}
				\For {$i \in 1,\ldots,p$}
					\State{$\nu \gets \frac{\lambda s}{p|\hat{\theta}^*_{t_i}|}$}
					\State{$\theta_{t_i}^l$ $\gets$ \Call{Prox}{$\theta_{t_i}^l - s\nabla f(\theta_t^l)_i, \nu$}}
				\EndFor	
			
			\State{$l \gets l+1$}
		\EndWhile
	
		\State{$D_t \gets \text{Diag}(\mathbf{1}(\hat{\theta}_{t_i}\neq0)|\hat{\theta}_{t_i}||\hat{\theta}_{t_i}^*| + (1-\mathbf{1}(\hat{\theta}_{t_i}\neq0))|\hat{\theta}_{t_i}^*|^2;i=1,\ldots,p)$}
		\State{$A_t \gets X_t^TW_t^{-1}X_t + \lambda D_t^{-1}+ \tau^* \Sigma^{-1}_{t|t-1}$}
		\State{$\Sigma_t \gets A_t^{-1}(X_t^TW_t^{-1}X_t + {\tau^*}^2\Sigma_{t|t-1}^{-1})A_t^{-1}$}
		\State \Return {$\hat{\theta}_{t}, \Sigma_t$}
	\EndFunction
	\Statex

\Initialize{
			$n_1 \gets \arg_{n_{1}} X_{1}\in\mathbb{R}^{n_{1}\times p}$\\
			$\hat{\theta}_0 \gets (X_1^TX_1)^{-1}X_1y_1$ \\
			\Statex{$w^2 \gets (y_1 - X_1\hat{\theta}_0)^T(y_1 - X_1\hat{\theta}_0)/(n_1 - 1)$}
			\Statex{$\Sigma_0 \gets I_p $}
			}
\Statex{\textbf{IRS Estimation for, } $t=1,2,\ldots$}
	\State{$n_t \gets \arg_{n_{t}} X_{t}\in\mathbb{R}^{n_{t}\times p}$}
	\State{$\hat{\theta}_{t|t-1} \gets F_t\hat{\theta}_{t-1}$}
	\State{$\Sigma_{t|t-1} \gets F_t\Sigma_{t-1}F_t^T + Q_t$}
	\State{$w^2 \gets (y_t - X_t\hat{\theta}_{t|t-1})^T(y_t - X_t\hat{\theta}_{t|t-1})/(n_t - 1) $}
	\State{$\hat{\theta}_t,\Sigma_t$ $\gets$ \\ \Call{ProximalDescent}{$X_t, y_t, \hat{\theta}_{t|t-1}, \Sigma_{t|t-1}^{-1}, w^2, \lambda, \tau$}}
	
\Statex	
	\algrule	
\OUTPUT {Model estimate: $\hat{\theta}_t, \Sigma_t$}

\end{algorithmic}

\caption{Inertial Regularization and Selection (IRS) estimation\label{alg:IRS-algo}}
\end{algorithm}

\subsection{Ability to handle missing values and structural changes}

A beneficial feature of IRS is the ease of handling missing values
and any change in model structure. A change in model structure implies
any addition or removal of model predictors. In case of missing values,
IRS can still provide a reasonable parameter estimate in a modeling
epoch by using the prior knowledge. 

If a new predictor variable is added, for eg., one new product at
Walmart, it can be easily incorporated in the modeling by expanding
$\hat{\theta}_{t|t-1}$ and $\hat{\Sigma}_{t|t-1}$. If there is no
expert prior knowledge for the new predictor, $\hat{\theta}_{t|t-1}$
can be expanded with 0 values and $\hat{\Sigma}_{t|t-1}$ with 0 off-diagonal
(no correlation) and a large variance. This is same as an uninformative
prior used is a Bayesian estimation. However, in several cases there
is some expert knowledge. For instance, in the above Walmart example
if the new product is a new baby diaper, we will have some knowledge
for its parameter from similar products, and/or its correlations with
other baby and related products.

Besides, IRS can handle removal of a predictor more cleverly than
most other conventional methods. A removed predictor does not mean
its effect will be gone immediately. In fact it will take some time
before its effect vanishes completely. For eg. suppose a TV brand,
say XYZ, is removed from a Walmart store, but since XYZ's sales must
be interacting with other TV sales, the effect of its absence will
be felt for some time (maybe raising the demand for others) before
gradually dying out. In IRS, it is not necessary to manually remove
any predictor variable from the model even if we know it is physically
removed. The inertial and selection component in IRS will automatically
update the resultant effect and eventually remove the predictor variables
from the model when its effect declines completely.

It is worthwhile to note that this IRS property also handles situations
when the removal of the predictor is temporary. In the above Walmart
eg., a reintroduction of the TV XYZ can be easily incorporated in
the modeling updates, with the effect of reintroduction being the
opposite of removal \textemdash{} the effect gradually increasing
in the model.

This ability to easily handle missing values and structural changes
in model is a valuable property because it makes the sequential modeling
more robust to the real world scenarios.

\section{Experimental Validation}
\label{sec:Experimental-Validation}

In this section, we experimentally validate the efficacy of IRS. We
compare it with Lasso, Kalman Filter (KF) and Ensemble KF (EnKF).
Lasso was applied locally on each epoch $t$to show a prediction error
baseline if the prior knowledge is ignored in sequential process.
KF and EnKF are commonly used state-of-the-art state-space models
against which IRS's performance will be gauged.

\begin{figure}
\begin{centering}
\begin{minipage}[t]{0.45\columnwidth}%
\subfloat[$p=500$\label{fig:simu-accu-1}]{\begin{centering}
\includegraphics[scale=0.15]{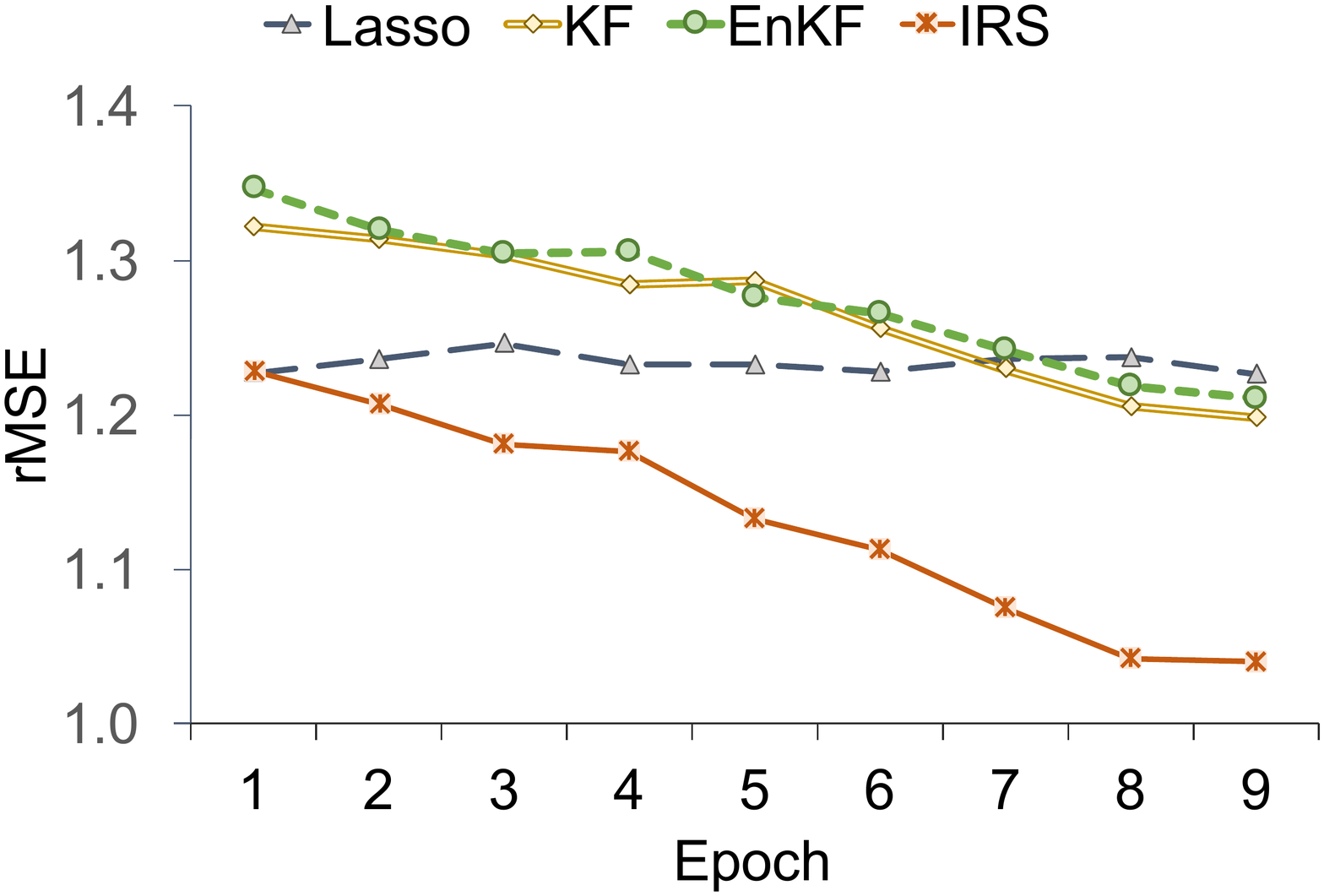}
\par\end{centering}
\centering{}}%
\end{minipage}\hspace{10bp}%
\begin{minipage}[t]{0.45\columnwidth}%
\subfloat[$p=5000$\label{fig:simu-accu-2}]{\begin{centering}
\includegraphics[scale=0.15]{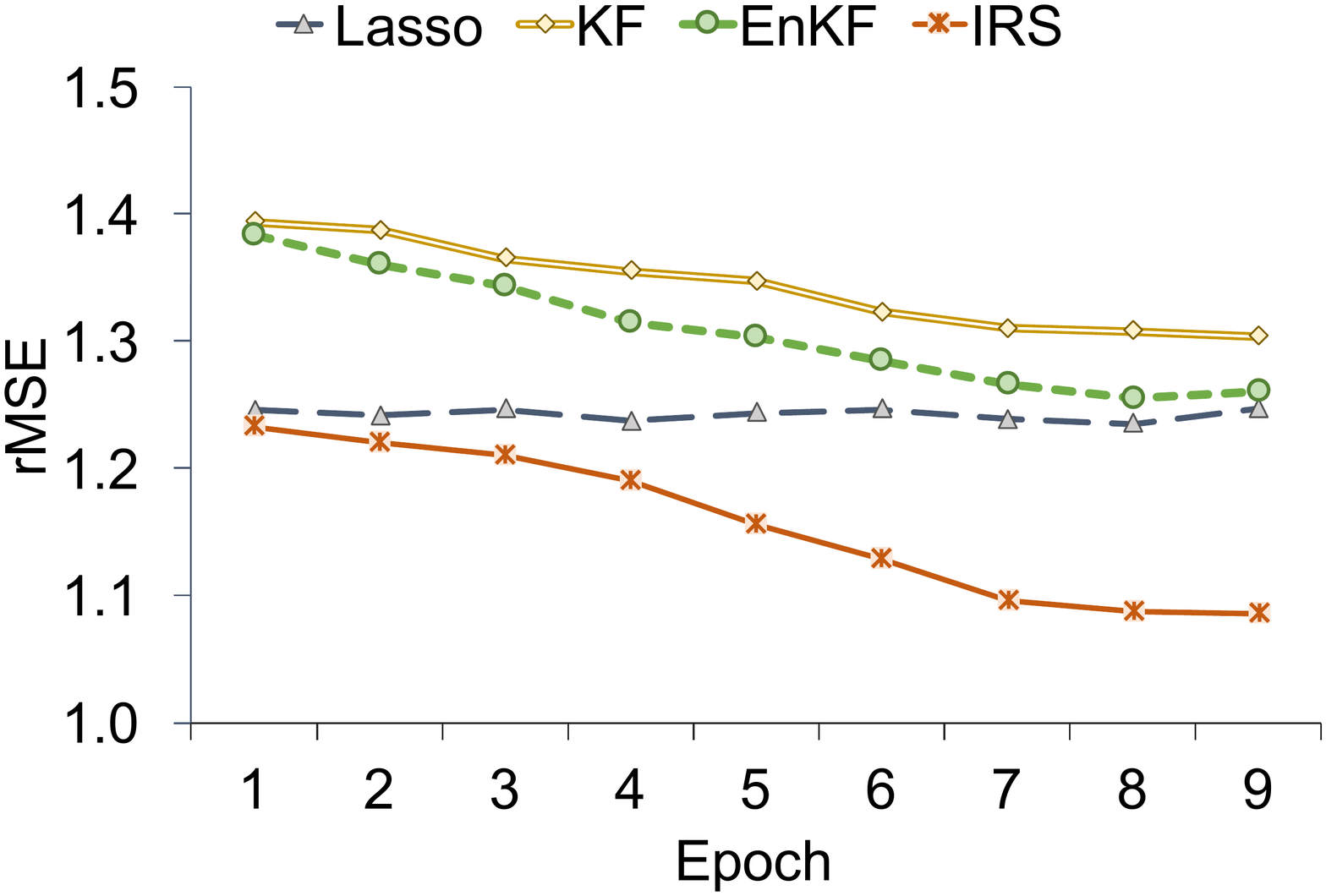}
\par\end{centering}
\centering{}}%
\end{minipage}
\par\end{centering}
\caption{Exp-1: Average rMSE of test errors from a 10-fold cross validation.\label{fig:Simu-Comparison-of-rMSE}}

\end{figure}

We setup two types of experiments for validation, denoted as Exp-1
and Exp-2. In Exp-1, the number of model parameters (predictors) are
kept same for all epochs, $p_{t}=p=\{500,5000\},t=1,2,\ldots,T$.
The model parameters are randomly initialized for $t=1$ from a multivariate
normal distribution with mean as 0, and an arbitrary covariance, such
that a sparsity level of 0.2 is maintained. As the model enters from
epoch $t$ to $t+1$, its parameters change with a random walk of
variance 1. The sample size is randomly selected between $(1.8p,2.1p)$.
The data is generated for 9 epochs, i.e. $T=9$. A 10-fold CV is performed
on initial three epochs sample for regularization parameter selection.
Thereafter, again a 10-fold CV is performed to fit each method, and
the average test prediction error, rMSEs, are shown in Fig.~\ref{fig:simu-accu-1}-\ref{fig:simu-accu-2}.

The figures show that Lasso has a similar rMSE as IRS in the initial
epoch. Thereafter, while Lasso's prediction error remain approximately
at the same level, IRS's error reduces. This is due to the use of
knowledge used from the past for updating the model with current data.
KF and EnKF similar had a reducing error rates, however, they had
a higher error than IRS and Lasso for the most of epochs. This can
be attributed to their inability to work in a high-dimensional sparse
processes. Besides, their performance seems to worsen when $p=5000$,
compared to when $p=500$. This maybe due to KF and EnKF overfitting
the data, resulting into higher test prediction errors, that gets
higher as dimension increases. 

In Exp-2, we initialized the model similar as Exp-1. However, for
subsequent epochs, $t>1$, a parameter in $\theta_{t}$ was changed
as, a) if $\theta_{t-1_{i}}$ is 0, it can become non-zero at $t$
with a small activation probability $(=0.05)$, and its value is again
drawn from the initial multivariate normal distribution, b) if $\theta_{t-1_{i}}\neq0$,
then it can be deactivated (made equal to $0$) if it is less than
a deactivation threshold $(=0.1)$ with a probability $(=0.1)$, c)
if $\theta_{t-1_{i}}\neq0$ and it is not deactivated, then it is
changed with a random walk, $F_{t_{i}}=1$, or directionally $F_{t_{i}}=(1+N(0,0.1))$
with added random noise $(\sim N(0,1))$.

This setup emulates a real world problem where, an otherwise insignificant
variable becomes significant and vice-versa, due to any systemic changes.
Also, the shift in $\theta$ is sometimes directional, indicating
an evolution. Regardless, we will perform the sequential modeling
assuming $F_{t}=I$, to stay in tandem with the common state-space
modeling practice.

In addition to incorporating evolution, we also reduce the data sample
size for progressing epochs to test each methods' efficacy in working
with growing prior knowledge and lesser data. 

\begin{figure}
\begin{centering}
\begin{minipage}[t]{0.45\columnwidth}%
\subfloat[$p=500$\label{fig:simu-accu-1-1}]{\begin{centering}
\includegraphics[scale=0.15]{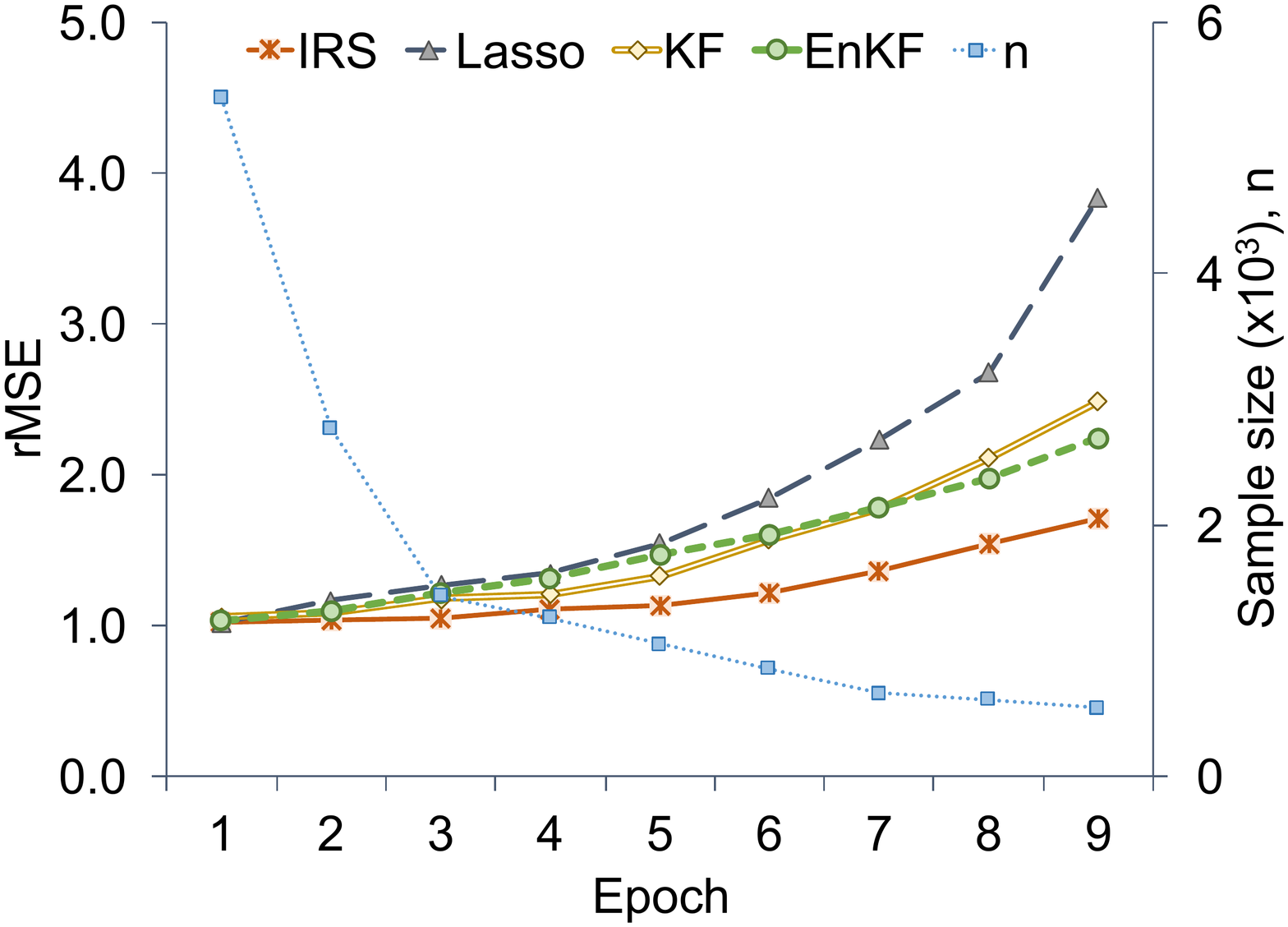}
\par\end{centering}
\centering{}}%
\end{minipage}\hspace{10bp}%
\begin{minipage}[t]{0.45\columnwidth}%
\subfloat[$p=5000$\label{fig:simu-accu-2-1}]{\begin{centering}
\includegraphics[scale=0.15]{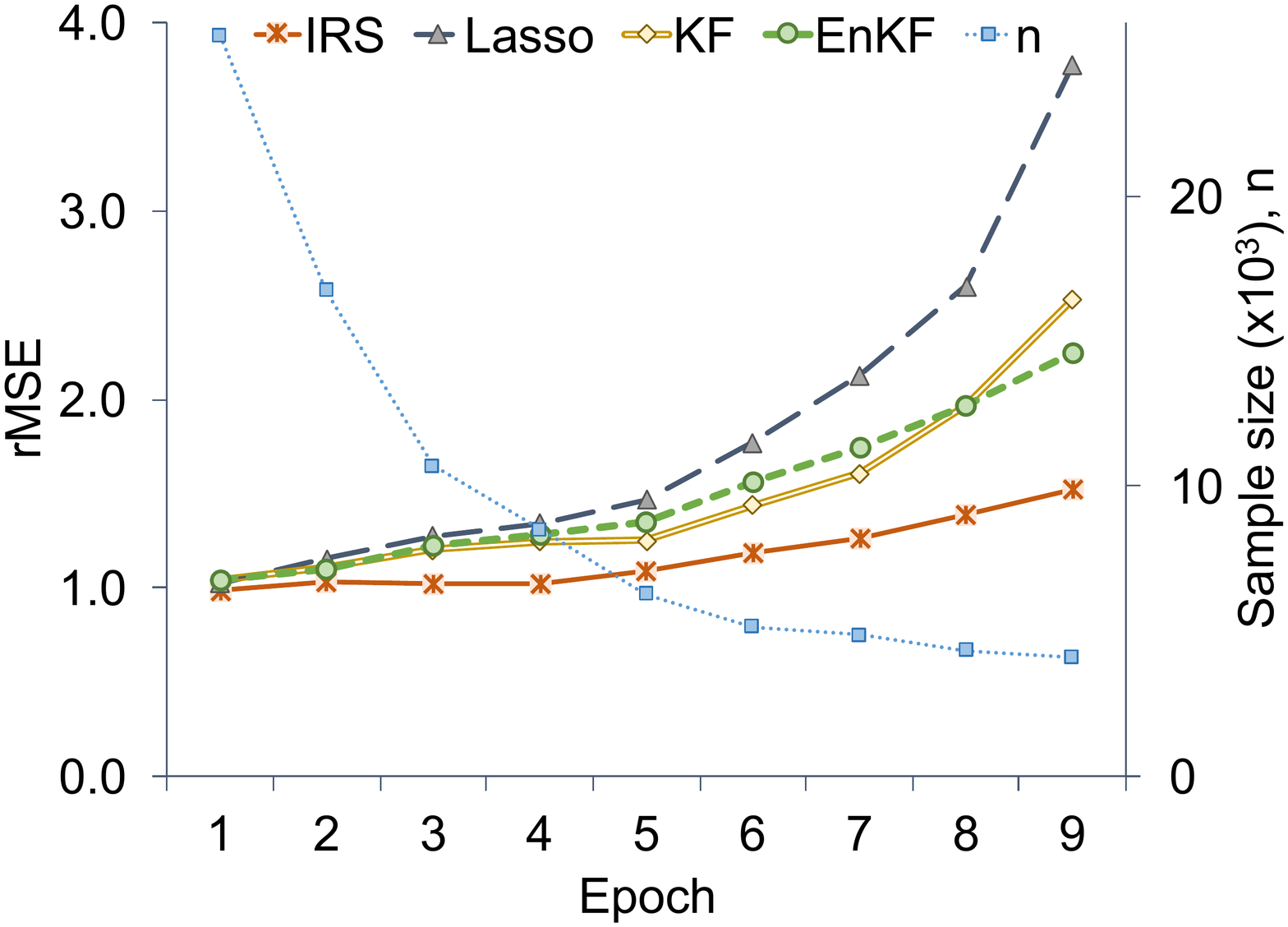}
\par\end{centering}
\centering{}}%
\end{minipage}
\par\end{centering}
\caption{Exp-2: Average rMSE of test errors from a 10-fold cross validation.\label{fig:Simu-Comparison-of-rMSE-1}}
\end{figure}

Again, a 10-fold CV is performed and the average rMSEs from test prediction
is reported in Fig.~\ref{fig:simu-accu-1-1}-\ref{fig:simu-accu-2-1}.
As expected, the errors increase with passing epochs due to lesser
sample size. While in the initial epochs, all methods have roughly
the same error rate, it increases sharply for a localized Lasso fit.
Between, EnKF and KF, the latters error rate grows faster and surpasses
the rMSE of EnKF after some epochs pass. This can be due to EnKF's
better ability to estimate the parameter covariance even with lesser
data (using the Monte Carlo setup). Among all, the increase in error
rate is the lowest for IRS. 

This result shows that IRS is also robust to the evolutionary model
changes, and can work with lesser data as it garners more prior knowledge.
This is an important feature for sequential modeling \textemdash{}
indicating the ability of quickly updating the model as the system
gets older.

The results in this section show the importance of use of past information
for sequentially modeling data that arrives in a streaming fashion.
For high-dimensional sparse and evolving models, it is shown that
IRS outperforms the state-space models popular for such sequential
modeling. Next, we will show a real world application of IRS.

\section{Application on an online Retail Data}
\label{sec:Case-Study}

Here we show a real world application of IRS using a retail data from
a UK-based and registered non-store online retail. The company mainly
sells unique all-occasion gifts to mostly wholesalers and some individuals.
The company was established in 1981 and relied on direct mailing catalogues.
It went online in about 2010, since when it started accumulating huge
amount of data. 

A subset of that data used in this study lies between Dec, 2010 to
Dec, 2011. The data is a customer transaction dataset, which has several
variables, namely, product name, quantity sold, invoice date-time,
unit price, customer-id, and country. For our analysis, we drop the
variable customer-id, and select data from the country, UK. From this
filtered data, we take a subset belonging to 41 products, such that
it comprises of high to low selling items.

The ``quantity sold'' for a product in a transaction is the response
variable, and the model's objective is to predict it using all available
predictors. For predictors, we extract the day-of-week and quarter-of-day
from the invoice date-time to form predictors. Month is not taken
because the data is for just one year, thus, deriving month seasonal
effects will not be possible. Besides, other predictors are the unit
price and product name. We create dummy columns for the categorical
product name and day-of-week variables. Finally, a predictor matrix
is made by taking in all predictor variables and their second-order
interactions. Interaction effects are important to consider because
sales of one product usually effects sales of some others. In total,
the predictor matrix has 363 columns, that is, we have a 363-dimensional
model.

\begin{figure}
\centering{}%
\begin{minipage}[t]{0.45\columnwidth}%
\includegraphics[scale=0.15]{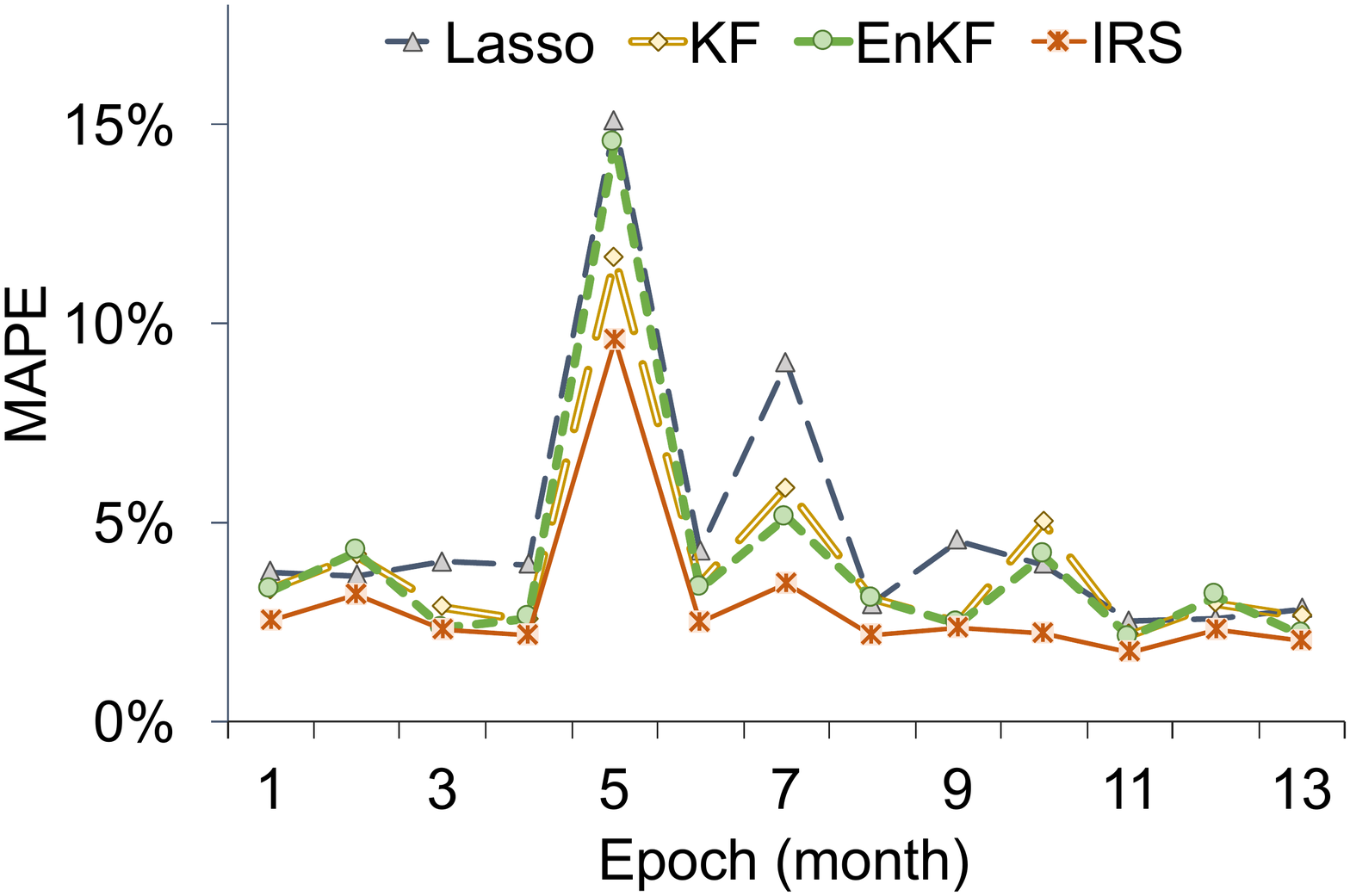}

\caption{Forecasting accuracy for a randomly selected set of items.\label{fig:real-data-1}}
\end{minipage}\hspace{10bp}%
\begin{minipage}[t]{0.45\columnwidth}%
\includegraphics[scale=0.15]{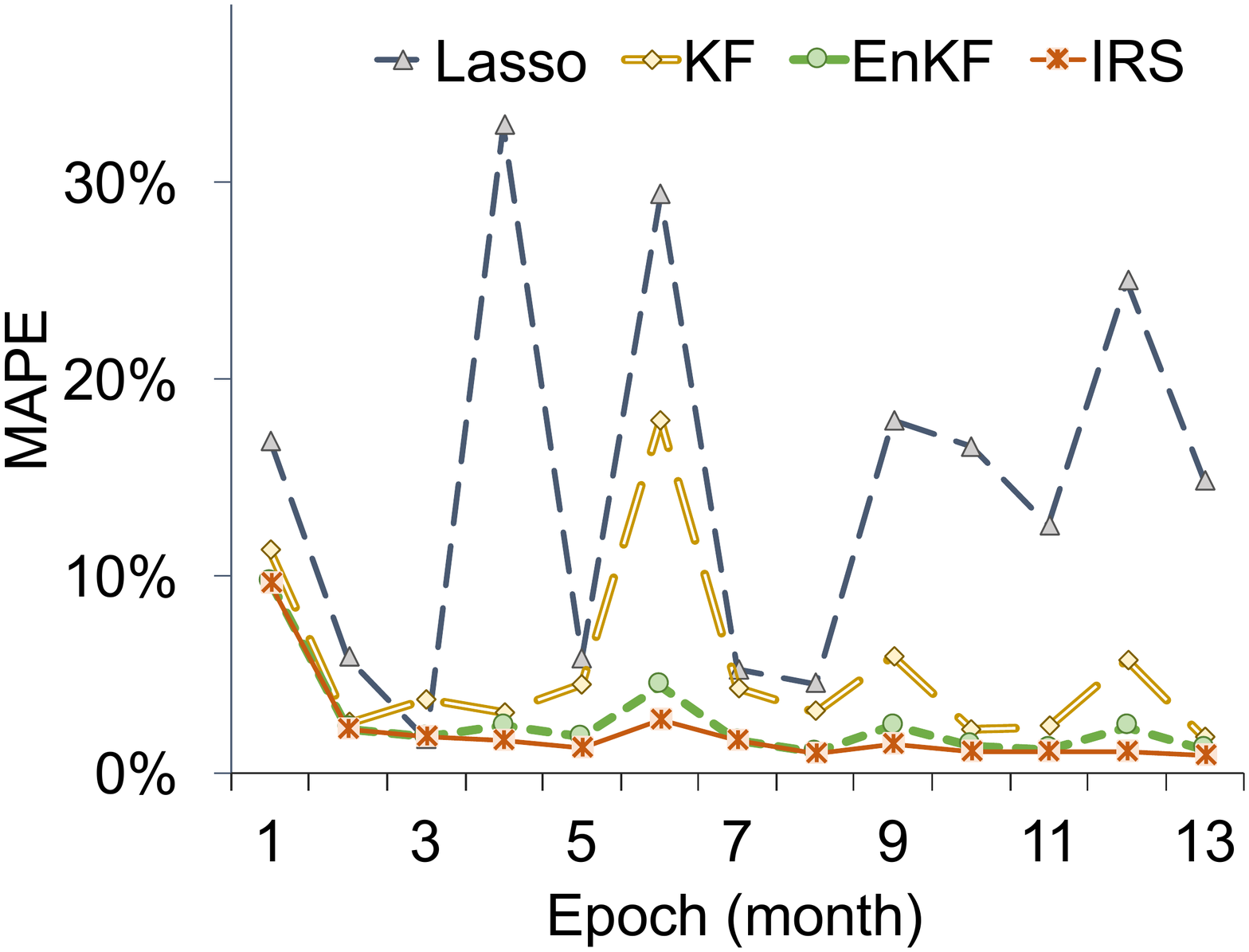}

\caption{Forecasting accuracy for low selling items.\label{fig:real-data-2}}
\end{minipage}
\end{figure}

Each month is treated as a modeling epoch. A 10-fold CV is performed
on the first three epochs for penalty parameters, $(\lambda,\tau)$,
selection, and then IRS is performed sequentially on all epochs. The
10-fold CV test prediction error (mean absolute percentage error,
MAPE) in Fig.~\ref{fig:real-data-1} show that IRS is more accurate
than others. This is because several predictors in the above model
will be statistically insignificant, thus, would require an effective
model selection, in addition to use of the prior knowledge. However,
as also mentioned before, KF and EnKF cannot effectively find a parsimonious
model, and, thus, has poorer test accuracy.

In addition, we show another result of prediction errors for low selling
products in Fig.~\ref{fig:real-data-2}. Demand forecast of low selling
items are usually more difficult because of fewer available data.
As shown in the figure, IRS again outperfoms other models. This can
be attributed to IRS's better capability is working with lesser data.

An online retail data, such as the one used here, is often very high-dimensional
due to several categorical variables and interactions. Besides, the
demand patterns typically have an inherent evolutionary characteristics.
Note that, this evolution is different from the seasonalities and
trend. Evolution is a gradual change due to various external factors,
that are hard to measure. IRS is, thus, well suited for such problems.
It can account for the evolutionary changes, and can handle high-dimensions,
as well as, an automatic model selection.

\section{Discussion and Conclusion}
\label{sec:Discussion-and-Conclusion}

In the previous section (\S\ref{sec:Case-Study}), we show IRS's
superior prediction accuracy in an online retail data problem. This
is an example of a high-dimensional statistical learning problem with
sparsity and large data collection in a streaming fashion, where the
process can also evolve. A majority of statistical learning problems
found these days have similar properties. For instance, sensors data,
telecommunications, weblogs, console logs, user music/movie play history
(on services like spotify and Netflix), etc. Besides, in such problems
various variables get added or removed from the process over time.
Under all this fast-paced dynamics, IRS can be easily applied and
yield a high accuracy.

In general, IRS's ability to auto-tune its regularization parameters
(given in \S\ref{subsec:Regularization-parameters}) from the
data makes it easy to implement. Due to this auto-tune ability, as
further elucidated in \S\ref{subsec:Algorithm}, IRS has just
two hyperparameters, $(\lambda,\tau)$, which if suitably selected
provides an optimal solution for any epoch $t=1,2,\ldots$. However,
in event of any abnormal or extreme system change, $(\lambda,\tau)$
may need to be re-tuned. Nevertheless, under normal conditions IRS
provides a seamless sequential estimation for a statistical learning
model.

Besides, the combination of inertial and $L_{1}$ penalty terms bring
important properties of \emph{resemblance }and \emph{model selection}.
Due to the inertia penalty, the model does not drastically change
between epochs, thus, keeps a resemblance with what an analyst already
knows about the model. The $L_{1}$ penalty provides an appropriate
model selection on top of it. It must be noted that in model selection,
the prior knowledge is also incorporated by using an adaptive penalty.
This prevents the $L_{1}$ penalty from selecting an extremely different
model than the past. 

However, a potential downside of the \emph{resemblance }property could
be if the initial model is severely sub-optimal, the subsequent models
can also be poor. In such cases, it may take several epochs for the
model estimate to get close to the optimal. Therefore, to avoid this,
it is recommended in \S\ref{subsec:Algorithm} to use a larger
amount of data in the initial epoch.

Regardless, IRS is also robust to noise and missing values due to
the inertia component. This, alongwith an optimal model selection
via $L_{1}$ penalty significantly boosts the test prediction accuracy,
demonstrated in \S\ref{sec:Experimental-Validation}-\ref{sec:Case-Study}.

In this paper, we show IRS's equivalence with Lasso, Kalman Filter
and Bayesian estimation in \S\ref{subsec:Lasso-equivalence}-\ref{subsec:Equivalent-Bayesian-Formulation}.
This also shows the interconnection between them for a statistical
learning. Specifically, Kalman filters are often seen as a different
family of estimators isolated from regularization techniques. Here,
we establish their connection, and also present a faster implementation
for Kalman filter in Appendix-D.

In summary, we develop a new method \textendash{} Inertial Regularization
and Selection (IRS) \textendash{} for sequentially modelin an evolving
statistical learning model. We present the modeling approach for a
linear state-change (evolution) and a linear state-response model
with gaussian errors. However, it can be easily extended to any non-linear
model, if there is a link function transforming it into a linear form.
For eg., a binary response in a classification model can be modeled
using a logit-link for the state response model. The method is shown
to outperform existing methods and several real world applications.

\bibliographystyle{abbrv}
\bibliography{sigproc}  % sigproc.bib is the name of the Bibliography in this case
% ACM needs 'a single self-contained file'!
%\balancecolumns
\vspace{-0.1in}
\appendix
%Appendix A
\section{Online recursive estimation loss function}
The parameter estimate is a Bayesian formulation can be found from
the Maximum a Posteriori (MAP) function, as shown below,

\begin{eqnarray}
\{\hat{\theta}_{t}\}_{t=\{1,\ldots T\}} & = & \arg\max\left[\left(\prod_{t=1}^{T}p(\theta_{t}|\theta_{t-1})p(y_{t}|\theta_{t}\right)\right.\nonumber \\
 &  & \left.p(y_{0}|\theta_{0})p(\theta_{0})\right]\nonumber \\
 & = & \arg\max\left[p(y_{T}|\theta_{T})p(\theta_{T}|\theta_{T-1})\right.\cdot\nonumber \\
 &  & \left(\prod_{t=1}^{T-1}p(\theta_{t}|\theta_{t-1})p(y_{t}|\theta_{t})\right)\nonumber \\
 &  & \left.p(y_{0}|\theta_{0})p(\theta_{0})\right]\label{eq:Apx-A-1}
\end{eqnarray}

A marginalization can be performed on Eq.~\ref{eq:Apx-A-1} to find
a globally optimal solution at the $T^{th}$ epoch.

\begin{eqnarray}
\hat{\theta}_{T} & = & \arg\max_{\theta_{T}}\left[\int_{\mathbb{R}^{T}}p(y_{T}|\theta_{T})p(\theta_{T}|\theta_{T-1})\left(\prod_{t=1}^{T-1}p(\theta_{t}|\theta_{t-1})\cdot\right.\right.\nonumber \\
 &  & \left.\left.p(y_{t}|\theta_{t})\right)\cdot p(y_{0}|\theta_{0})p(\theta_{0})d\{\theta_{i}\}_{i=1,\ldots,T-1}\right]\nonumber \\
 & = & \arg\max\left[p(y_{T}|\theta_{T})p(\theta_{T}|\theta_{T-1})\int_{\mathbb{R}^{T}}\left(\prod_{t=1}^{T-1}p(\theta_{t}|\theta_{t-1})\cdot\right.\right.\nonumber \\
 &  & \left.\left.p(y_{t}|\theta_{t})\right)p(y_{0}|\theta_{0})p(\theta_{0})d\{\theta_{i}\}_{i=1,\ldots,T-1}\right]\label{eq:Apx-A-2}
\end{eqnarray}

The integral part in Eq.~\ref{eq:Apx-A-2} is essentially the prior
on $\theta_{T}$. Thus, for an online estimation, the updated form
of Eq.~\ref{eq:Apx-A-1} for $\theta_{T}$, given below, can be used.

\begin{eqnarray}
\hat{\theta}_{T} & = & \arg\max_{\theta_{T}}p(y_{T}|\theta_{T})p_{\hat{\theta}_{T-1}}(\theta_{T})\label{eq:Apx-A-3}
\end{eqnarray}

Eq.~\ref{eq:Apx-A-3} summarizes all the past information from the
likelihood $\theta_{T}|\{y_{t}\}_{t=1,\ldots,T-1}$ into the prior
for $\theta_{T}$ given $\hat{\theta}_{T-1}$. Using the Bayesian
prior distribution for IRS given in \S\ref{subsec:Equivalent-Bayesian-Formulation},
into Eq.~\ref{eq:Apx-A-3}, we get,

\begin{eqnarray*}
\hat{\theta}_{T} & = & \arg\max_{\theta_{T}}\exp\left[-(y_{T}-X_{T}\theta_{T})^{T}W_{T}^{-1}(y_{T}-X_{T}\theta_{T})\right]\cdot\\
 &  & \exp\left[-\tau(\theta_{T}-\hat{\theta}_{T|T-1})^{T}\Sigma_{T|T-1}^{-1}(\theta_{T}-\hat{\theta}_{T|T-1})-\right.\\
 &  & \left.\sum\lambda_{t_{i}}\frac{|\theta_{T_{i}}|}{|\hat{\theta}_{T_{i}}^{*}|}\right]\\
 & = & \arg\min_{\theta_{T}}(y_{T}-X_{T}\theta_{T})^{T}W_{T}^{-1}(y_{T}-X_{T}\theta_{T})+\\
 &  & \tau(\theta_{T}-\hat{\theta}_{T|T-1})^{T}\Sigma_{T|T-1}^{-1}(\theta_{T}-\hat{\theta}_{T|T-1})+\sum\lambda_{t_{i}}\frac{|\theta_{T_{i}}|}{|\hat{\theta}_{T_{i}}^{*}|}
\end{eqnarray*}

% This next section command marks the start of
% Appendix B, and does not continue the present hierarchy
\section{OLS estimate for the LASSO equivalent}

The OLS estimate, denoted by $\hat{\theta}_{t}^{*}$ , for IRS can
be easily found from its lasso equivalent given in Eq.~\ref{eq:Lasso-equi},
in \S\ref{subsec:Lasso-equivalence}.

\begin{eqnarray}
\hat{\theta}_{t}^{*} & = & (\tilde{X}_{t}^{T}\tilde{X}_{t})^{-1}\tilde{X}_{t}^{T}\tilde{y}_{t}\nonumber \\
 & = & \left(\frac{1}{2n_{t}}X_{t}^{T}W_{t}^{-1}X_{t}+\frac{\tau}{2p_{t}}\Sigma_{t|t-1}^{-1}\right)^{-1}\cdot\nonumber \\
 &  & \left(\frac{1}{2n_{t}}X_{t}^{T}W_{t}^{-1}y_{t}+\frac{\tau}{2p_{t}}\Sigma_{t|t-1}^{-1}\hat{\theta}_{t|t-1}\right)\nonumber \\
 & = & \left(X_{t}^{T}W_{t}^{-1}X_{t}+\tau^{*}\Sigma_{t|t-1}^{-1}\right)^{-1}\left(X_{t}^{T}W_{t}^{-1}y_{t}+\tau^{*}\Sigma_{t|t-1}^{-1}\hat{\theta}_{t|t-1}\right)\label{eq:OLS-lasso}
\end{eqnarray}

where, $\tau^{*}=\frac{\tau n_{t}}{p_{t}}$.

\section{Proving LASSO equivalence}
Using an augmented $X$ and $y$ in Eq.~\ref{eq:Xy-augmentation},
we need to show,

\begin{eqnarray*}
 &  & (\tilde{y}_{t}-\tilde{X}_{t}\theta_{t})^{T}(\tilde{y}_{t}-\tilde{X}_{t}\theta_{t})_{\text{L.H.S.}}\\
 & = & \left[\frac{1}{2n_{t}}(Y_{t}-X_{t}\theta_{t})^{T}W_{t}^{-1}(Y_{t}-X_{t}\theta_{t})+\right.\\
 &  & \left.\frac{\tau}{2p_{t}}(\theta_{t}-\hat{\theta}_{t|t-1})^{T}\Sigma_{t|t-1}^{-1}(\theta_{t}-\hat{\theta}_{t|t-1})\right]_{\text{R.H.S.}}
\end{eqnarray*}
\vspace{-0.2in}

To prove L.H.S. is equal to R.H.S., we expand L.H.S.:

\begin{eqnarray*}
 &  & (\tilde{y}_{t}-\tilde{X}_{t}\theta_{t})^{T}(\tilde{y}_{t}-\tilde{X}_{t}\theta_{t})\\
 & = & \tilde{y}_{t}^{T}\tilde{y}_{t}-2\tilde{y}_{t}^{T}\tilde{X}_{t}\theta_{t}+\theta_{t}^{T}\tilde{X}_{t}^{T}\tilde{X}_{t}\theta_{t}\\
 & = & \frac{1}{2n_{t}}y_{t}^{T}W_{t}^{-1}y_{t}+\frac{\tau}{2p_{t}}\hat{\theta}_{t|t-1}^{T}\Sigma_{t|t-1}^{-1}\hat{\theta}_{t|t-1}\\
 &  & -2\left(\frac{1}{2n_{t}}y_{t}^{T}W_{t}^{-1}X_{t}\theta_{t}+\frac{\tau}{2p_{t}}\hat{\theta}_{t|t-1}^{T}\Sigma_{t|t-1}^{-1}\theta_{t}\right)\\
 &  & +\frac{1}{2n_{t}}\theta_{t}^{T}X_{t}^{T}W_{t}^{-1}X_{t}\theta_{t}+\frac{\tau}{2p_{t}}\theta_{t}^{T}\Sigma_{t|t-1}^{-1}\theta_{t}\\
 & = & \frac{1}{2n_{t}}\left(y_{t}^{T}W_{t}^{-1}y_{t}-2y_{t}^{T}W_{t}^{-1}X_{t}\theta_{t}+\theta_{t}^{T}X_{t}^{T}W_{t}^{-1}X_{t}\theta_{t}\right)\\
 &  & +\frac{\tau}{2p_{t}}\left(\theta_{t}^{T}\Sigma_{t|t-1}^{-1}\theta_{t}-2\hat{\theta}_{t|t-1}^{T}\Sigma_{t|t-1}^{-1}\theta_{t}+\hat{\theta}_{t|t-1}^{T}\Sigma_{t|t-1}^{-1}\hat{\theta}_{t|t-1}\right)\\
 & = & \frac{1}{2n_{t}}(y_{t}-X_{t}\theta_{t})^{T}W_{t}^{-1}(y_{t}-X_{t}\theta_{t})+\\
 &  & \frac{\tau}{2p_{t}}(\theta_{t}-\hat{\theta}_{t|t-1})^{T}\Sigma_{t|t-1}^{-1}(\theta_{t}-\hat{\theta}_{t|t-1})\\
 & = & \text{R.H.S.}
\end{eqnarray*}
\vspace{-0.2in}

\section{Showing Kalman Filter Equivalence}

Here we expand the IRS expression for $\theta_{t}$ estimate when
$\lambda=0$, denoted by $\hat{\theta}_{t}^{*}$, given in Eq.~\ref{eq:OLS-lasso}
in Appendix-B.
\vspace{-0.2in}
\begin{eqnarray}
\hat{\theta}_{t}^{*} & = & \left[X_{t}^{T}W_{t}^{-1}X_{t}+\tau^{*}\Sigma_{t|t-1}^{-1}\right]^{-1}\cdot\nonumber \\
 &  & \left[X_{t}^{T}W_{t}^{-1}y_{t}+\tau^{*}\Sigma_{t|t-1}^{-1}\hat{\theta}_{t|t-1}\right]\nonumber \\
 & = & \left[\frac{1}{\tau^{*}}\Sigma_{t|t-1}-\frac{1}{\tau^{*}}K_{t}X_{t}\frac{1}{\tau^{*}}\Sigma_{t|t-1}\right]\cdot\nonumber \\
 &  & \left[X_{t}^{T}W_{t}^{-1}y_{t}+\tau^{*}\Sigma_{t|t-1}^{-1}\hat{\theta}_{t|t-1}\right]\nonumber \\
 & = & \hat{\theta}_{t|t-1}-\frac{1}{\tau^{*}}K_{t}X_{t}\hat{\theta}_{t|t-1}+\nonumber \\
 &  & \left[\frac{1}{\tau^{*}}\Sigma_{t|t-1}X_{t}^{T}W_{t}^{-1}-\frac{1}{(\tau^{*})^{2}}K_{t}X_{t}\Sigma_{t|t-1}X_{t}^{T}W_{t}^{-1}\right]y_{t}\nonumber \\
 & = & \hat{\theta}_{t|t-1}-\frac{1}{\tau^{*}}K_{t}X_{t}\hat{\theta}_{t|t-1}+\nonumber \\
 &  & K_{t}\left[\frac{1}{\tau^{*}}\left(W_{t}+\frac{1}{\tau^{*}}X_{t}\Sigma_{t|t-1}X_{t}^{T}\right)W_{t}^{-1}-\right.\nonumber \\
 &  & \left.\frac{1}{(\tau^{*})^{2}}X_{t}\Sigma_{t|t-1}X_{t}^{T}W_{t}^{-1}\right]y_{t}\nonumber \\
 & = & \hat{\theta}_{t|t-1}+\frac{1}{\tau^{*}}K_{t}(y_{t}-X_{t}\hat{\theta}_{t|t-1})\label{eq:kalman-1-1}
\end{eqnarray}

where, $K_{t}=\Sigma_{t|t-1}X_{t}^{T}\left(W_{t}+\frac{1}{\tau^{*}}X_{t}\Sigma_{t|t-1}X_{t}^{T}\right)^{-1}$
, also known as Kalman gain. 

Besides, when $\tau^{*}=1$, the covariance of $\hat{\theta}_{t}^{*}$
can be shown to be same as in Kalman filter.

\begin{eqnarray*}
\Sigma_{t}^{*} & = & \text{cov}\left\{ \underbrace{\left[X_{t}^{T}W_{t}^{-1}X_{t}+\Sigma_{t|t-1}^{-1}\right]}_{E_{t}}^{-1}\cdot\right.\\
 &  & \left.\left[X_{t}^{T}W_{t}^{-1}y_{t}+\Sigma_{t|t-1}^{-1}\hat{\theta}_{t|t-1}\right]\right\} \\
 & = & E_{t}^{-1}X_{t}^{T}W_{t}^{-1}\text{cov}(y_{t})W_{t}^{-1}X_{t}^{T}E_{t}^{-1}+\\
 &  & E_{t}^{-1}\Sigma_{t|t-1}^{-1}\text{\ensuremath{\hat{\theta}_{t|t-1}}}\Sigma_{t|t-1}^{-1}E_{t}^{-1}\\
 & = & E_{t}^{-1}\left(X_{t}^{T}W_{t}^{-1}X_{t}^{T}+\Sigma_{t|t-1}^{-1}\right)E_{t}^{-1}\\
 & = & E_{t}^{-1}E_{t}E_{t}^{-1}\\
 & = & \left(X_{t}^{T}W_{t}^{-1}X_{t}^{T}+\Sigma_{t|t-1}^{-1}\right)^{-1}\\
 & = & \Sigma_{t|t-1}-\\
 &  & \Sigma_{t|t-1}X_{t}^{T}\left(W_{t}+X_{t}\Sigma_{t|t-1}X_{t}^{T}\right)^{-1}X_{t}\Sigma_{t|t-1}\\
 & = & \Sigma_{t|t-1}-K_{t}X_{t}\Sigma_{t|t-1}\\
 & = & (I-K_{t}X_{t})\Sigma_{t|t-1}
\end{eqnarray*}

\textbf{Faster Kalman filter algorithm}

Given a previous estimate of $\theta$ and its covariance, as $\hat{\theta}_{t-1}$
and $\Sigma_{t-1}$, and the data, the updated model is found using
the following steps,

\begin{eqnarray}
\hat{\theta}_{t|t-1} & = & F_{t}\hat{\theta}_{t-1}\\
\Sigma_{t|t-1} & = & F_{t}\Sigma_{t-1}F_{t}^{T}+Q_{t}\\
K_{t} & = & \Sigma_{t|t-1}X_{t}^{T}\left(\Sigma_{t}+X_{t}\Sigma_{t|t-1}X_{t}^{T}\right)^{-1}\label{eq:Kalman-Kt}\\
\hat{\theta}_{t} & = & \hat{\theta}_{t|t-1}+K_{t}\left(y_{t}-X_{t}\hat{\theta}_{t|t-1}\right)\label{eq:kalman-slow-theta}\\
\Sigma_{t} & = & \left(I-K_{t}X_{t}\right)\Sigma_{t|t-1}\label{eq:kalman-slow-sigma}
\end{eqnarray}

The formulation shows an evolving model as an additive function of
the prior model. However, computationally it becomes intensive if
the datasize is even reasonably large. This is due to a $n_{t}\times n_{t}$
matrix inversion required in Eq.~\ref{eq:Kalman-Kt}. Using the IRS
equivalence for Kalman, we can have the following procedure for Kalman
filter estimation, which is significantly faster when $p<n_{t}$,
i.e. the number of parameters being less than the datasize, which
is more common in statistical learning models (unlike the signal processing
model, a primary focus for Kalman filters).

A faster algorithm for Kalman will, thus, replace Eq.~\ref{eq:Kalman-Kt}-\ref{eq:kalman-slow-sigma}
as,

\begin{eqnarray*}
\hat{\theta}_{t} & = & \left[X_{t}^{T}R_{t}^{-1}X_{t}+\Sigma_{t|t-1}^{-1}\right]^{-1}\left[X_{t}^{T}R_{t}^{-1}y_{t}+\Sigma_{t|t-1}^{-1}\hat{\theta}_{t|t-1}\right]\\
\Sigma_{t} & = & \left(X_{t}^{T}R_{t}^{-1}X_{t}+\Sigma_{t|t-1}^{-1}\right)^{-1}
\end{eqnarray*}

\section{Stationarity of IRS loss function}
The IRS loss function given in Eq.~\ref{eq:main-loss-main-3} has
three components. Here we will show that they are individually stationary,
meaning they have a constant expectation, independent of time epochs.

\begin{eqnarray}
\mathcal{L}_{\hat{\theta}_{t-1}}(\theta_{t}) & = & \underbrace{\frac{1}{2n_{t}}(y_{t}-X_{t}\theta_{t})^{T}W_{t}^{-1}(y_{t}-X_{t}\theta_{t})}_{(\text{A})}+\nonumber \\
 &  & \underbrace{\frac{\tau}{2p_{t}}(\theta_{t}-\hat{\theta}_{t|t-1})^{T}\Sigma_{t|t-1}^{-1}(\theta_{t}-\hat{\theta}_{t|t-1})}_{(\text{B})}+\nonumber \\
 &  & \underbrace{\frac{\lambda}{p_{t}}\sum_{i=1}^{p_{t}}\frac{|\theta_{t_{i}}|}{|\hat{\theta}_{t_{i}}^{*}|}}_{(\text{C})}\label{eq:main-loss-2-1-1}
\end{eqnarray}

We will use the following identity applicable for a random variable,
$x\sim N(\mu,\Sigma)$, and any known matrix $A$.

\begin{eqnarray}
E[x^{T}Ax] & = & \text{Tr}(A\Sigma)+\mu^{T}A\mu\label{eq:stationary-id-1}
\end{eqnarray}

\begin{itemize}
\item Part $(\text{A})$: $y_{t}-X_{t}\theta_{t}\sim N(0,W_{t})$, where
$W_{t}$ is a symmetric matrix. Therefore, $E[(\text{A})]=\frac{1}{2n_{t}}\left(\text{Tr}(W_{t}^{-1}W_{t})+0\right)=\frac{1}{2n_{t}}n_{t}=\frac{1}{2}$.
\item Part $(\text{B})$: $\theta_{t}-\hat{\theta}_{t|t-1}\sim N(0,\Sigma_{t|t-1})$.
Similarly, as for part $(\text{A})$, we will have, $E[(\text{B})]=\frac{\tau}{2}$.
\item Part $(\text{C})$: Suppose, $\theta_{t_{i}}$ is approximated as
a hard-thresholded estimate from OLS, $\theta_{t_{i}}=\hat{\theta}_{t_{i}}^{*}\cdot\epsilon_{t_{i}}^{*}$,
where $\epsilon_{t_{i}}^{*}\sim\text{Bernoulli}(\alpha)$. Therefore,
$E\left[\left|\frac{\theta_{t_{i}}}{\hat{\theta}_{t_{i}}^{*}}\right|\right]=\alpha\implies E[(\text{C})]=\lambda\alpha$.
Here, $\alpha$ can be interpreted as the proportion of variables
that are selected from the $L_{1}$ penalization. Assuming $\alpha$
is same for all epochs, the expectation of part $(\text{C})$ is independent
of $t$.
\end{itemize}

\end{document}